\pdfoutput=1

\documentclass[12pt,amsmath,amssymb,longbibliography,nofootinbib,showkeys,eprint]{revtex4-2}

\usepackage[activate={true,nocompatibility},final,tracking=true,kerning=true,spacing=true]{microtype}
\usepackage{epigraph}
\usepackage{graphicx}
\usepackage{epstopdf}
\usepackage{braket}
\usepackage{bm}
\usepackage{numprint}
\usepackage{float}
\usepackage[T2A,T1]{fontenc}
\usepackage[utf8]{inputenc}
\usepackage{hhline}
\usepackage{xfrac}
\usepackage{amsthm}
\usepackage{makecell}
\usepackage{dsfont}
\usepackage{calc}
\usepackage{seqsplit}
\usepackage{hyperref}
\hypersetup{
    colorlinks=true
}
\usepackage{orcidlink}

\begin{document}
\preprint{V.M.}
\title{Partially Unitary Learning}

\author{Mikhail Gennadievich \surname{Belov}}
\email{mikhail.belov@tafs.pro}
\affiliation{Lomonosov Moscow State University,  Faculty of Mechanics and Mathematics,
   GSP-1,  Moscow, Vorob'evy Gory, 119991, Russia}

\author{Vladislav Gennadievich \surname{Malyshkin}\,\orcidlink{0000-0003-0429-3456}}
\email{malyshki@ton.ioffe.ru}
\affiliation{Ioffe Institute, Politekhnicheskaya 26, St Petersburg, 194021, Russia}

\date{May, 14, 2024}

\begin{abstract}
\begin{verbatim}
$Id: PartiallyUnitaryLearning.tex,v 1.400 2024/11/14 19:45:25 mal Exp $
\end{verbatim}
The problem of an optimal mapping 
between Hilbert spaces
\emph{IN} of
$\left|\psi\right\rangle$
and
\emph{OUT} of
$\left|\phi\right\rangle$
based on a set of wavefunction measurements (within a phase)
$\psi_l \to \phi_l$, $l=1\dots M$,
is formulated as an optimization problem maximizing the total fidelity
$\sum_{l=1}^{M} \omega^{(l)}
\left|\langle\phi_l|\mathcal{U}|\psi_l\rangle\right|^2$
subject to probability preservation constraints
on  $\mathcal{U}$ (partial unitarity).
The constructed operator
$\mathcal{U}$ can be considered as an \emph{IN} to \emph{OUT}
quantum channel;
it is a partially unitary rectangular matrix (an isometry) of dimension
$\dim(\emph{OUT}) \times \dim(\emph{IN})$ transforming
operators  as
$A^{\emph{OUT}}=\mathcal{U} A^{\emph{IN}} \mathcal{U}^{\dagger}$.
An iterative algorithm for finding the global maximum of this optimization problem is developed, and its application to a number of problems is demonstrated.
A software product implementing the algorithm
\href{http://www.ioffe.ru/LNEPS/malyshkin/code_polynomials_quadratures.zip}{is available}
from the authors.

\end{abstract}

\maketitle
\newpage

\noindent Dedicated to Professor Arthur McGurn on the occasion of his 75th birthday.

\section{\label{intro}Introduction}

Progress in machine learning (ML) knowledge representation,
from linear regression coefficients,
perceptron weights\cite{rosenblatt1958perceptron},
statistical learning\cite{vapnik1974method},
and logical approaches\cite{hajek1977generation}
to support vector machines\cite{vapnik2013nature},
rules and decision trees\cite{witten2002data},
fuzzy logic\cite{zadeh1965fuzzy,hajek1995fuzzy},
and deep learning\cite{bengio2013representation}
has defined the direction of ML development over the last four decades.
Recently, knowledge representation in the form of a unitary operator has started to attract significant attention\cite{bisio2010optimal,arjovsky2016unitary,hyland2017learning}.
The problem of learning unitary matrices is also useful in various other fields.
For example it can be applied to
the quantum mechanics inverse problem\cite{razavy2020introduction,johansson2012qutip},
investigating the dynamics of quantum many-body systems\cite{carleo2017solving,choi2016exploring,yan2017equilibration,bordia2017probing,yan2017dynamics},
quantum computing\cite{ramezani2020machine,kiani2020quantum,huang2021learning,torlai2023quantum},
light coherence\cite{pai2019matrix},
market dynamics\cite{malyshkin2022market}
and other fields.

The techniques used for unitary learning differ in unitary matrix representation, input data, and quality criteria.
A substantial number of existing works\cite{pai2019matrix,kiani2020learning,kiani2022projunn}
use the Frobenius $L^2$
norm of the difference between the target and current matrix, 
 $\left\|\mathcal{U}-\mathcal{V}\right\|^2$.
The main advantage of this approach is the applicability of first-order gradient optimization,
but it has all the limitations of $L^2$ minimization approaches.
A better option is to use the
\href{https://en.wikipedia.org/wiki/Fidelity_of_quantum_states}{fidelity}
of the target states $\left|\Braket{\phi|\mathcal{U}|\psi}\right|^2$; this approach is
utilized in \cite{bisio2010optimal,lloyd2019efficient} and many others.
An important advantage of this approach is that the multiplication of the source 
 $\Ket{\psi}$
or target $\Ket{\phi}$ by a random phase does not change the fidelity.
A disadvantage is that the operator $\mathcal{U}$ itself can only be determined within a phase.
There is a very interesting approach to representing a unitary matrix of dimension $n$
in a recursive way by splitting it into $[n/2]$-sized matrices and continuing the process recursively \cite{wang2024variational}.
This includes a unitary representation $\mathcal{U} \approx V_1 (U_A \otimes U_B) V_0$,
which requires the introduction of pre- and post-processing operators $V_1$ and $V_0$,
allowing for the identification of an appropriate cost function.

The algorithm developed in this paper is applicable only when the objective function is a quadratic function of
$\mathcal{U}$, as in the form given by the fidelity (\ref{allProjUKxfAppendix}) or the cost function (\ref{costVQA}) below.
This form allows the optimization problem to be formulated as a novel algebraic problem (\ref{eigenvaluesLikeProblem})
and avoids the difficult challenge of unitary matrix parametrization.
In our previous work\cite{malyshkin2022machine} the problem of unitary learning
was generalized to partially unitary operators.
This operator maps two Hilbert spaces of different dimensions, whereas a unitary operator maps a Hilbert space into itself.
The problem is to maximize the fidelity of a mapping
between the Hilbert spaces
\emph{IN} of
$\Ket{ \psi}$
and
\emph{OUT} of
$\Ket{\phi}$
based on a set of wavefunction measurement (within a phase) observations
$\psi_l \to \phi_l$, $l=1\dots M$,
as an optimization problem maximizing the total fidelity
$\sum_{l=1}^{M} \omega^{(l)}
\left|\Braket{\phi_l|\mathcal{U}|\psi_l}\right|^2$
subject to probability preservation constraints
on  $\mathcal{U}$ (partial unitarity).
This problem is reduced to a problem of maximizing a quadratic form on
$\mathcal{U}$
subject to quadratic form constraints.
This is a variant of the \href{https://en.wikipedia.org/wiki/Quadratically_constrained_quadratic_program}{QCQP} (Quadratically Constrained Quadratic Program) problem.
This problem is non-convex, exhibiting local extrema and multiple saddle points.
In this work, an iterative algorithm for finding the global maximum of this optimization problem is developed, and its application to a number of problems is demonstrated.

The quantum mechanics inverse problem of reconstructing $\mathcal{U}$ from measured wavefunction observations $\psi_l \to \phi_l$, $l=1\dots M$
is transformed into a new algebraic problem (\ref{eigenvaluesLikeProblem}).
This problem is analogous to the  Schr\"{o}dinger equation, where instead of a Hamiltonian, there is a superoperator $S$,
the ``eigenvector'' $\mathcal{U}$ corresponds to a unitary operator,
and the ``eigenvalue'' $\lambda$ corresponds to a Hermitian matrix.
The solution to the quantum mechanics inverse problem can be found by solving this  Schr\"{o}dinger-like equation.
This represents the most important result of the study.
Currently, only a numerical solution is available. 
For software availability, please refer to Appendix \ref{Software};
all references to code in the paper correspond to this software.

\section{\label{formulationOfTheProblem}Formulation of the Problem}
Consider the quantum mechanics inverse problem.
It may be broadly described as a problem of
determining the internal structure (e.g. Hamiltonian)
of a system from wavefunction measurements.
A number of other problems in statistics, machine learning, data analysis, etc., can be converted to a problem of the following form.
Let there be two Hilbert spaces of dimensions $n$ and $D$,
corresponding to states $\psi(\mathbf{x})$ and $\phi(\mathbf{f})$
in some $\mathbf{x}$- and $\mathbf{f}$- bases $x_k$ and $f_j$
where  $D\le n$
\begin{subequations}
\label{baseDefXF}
\begin{align}
\psi(\mathbf{x})&=\sum\limits_{k=0}^{n-1}\alpha_k x_k \label{baseX} \\
\phi(\mathbf{f})&=\sum\limits_{j=0}^{D-1}\beta_j f_j \label{baseF}
\end{align}
\end{subequations}
There is a scalar product operation $\Braket{\cdot}$ in each
Hilbert space allowing to calculate scalar product inside the basis
$\Braket{\psi|\psi^{\prime}}=\sum_{k,q=0}^{n-1}
\alpha^*_k
\Braket{x_k|x_q}
\alpha^{\prime}_q$
and $\Braket{\phi|\phi^{\prime}}=\sum_{j,i=0}^{D-1}
\beta^*_j
\Braket{f_j|f_i}
\beta^{\prime}_i$,
but \textsl{not across the bases}:
the $\Braket{x_k|f_j}$ cannot be calculated!
Assume we have $l=1\dots M$  wavefunction pairs (typically $M\gg n$)
as ``observations'':
\begin{align}
\psi_l(\mathbf{x})& \to \phi_l(\mathbf{f})
  & \text{weight $\omega^{(l)}$}  \label{mlproblemVector} \\
 1&=\Braket{\psi_l|\psi_l}=\Braket{\phi_l|\phi_l} \label{mlproblemVectorNorm}
\end{align}
These represent $M$ mappings from
Hilbert space $\mathbf{x}$ to Hilbert space $\mathbf{f}$.
The weights $\omega^{(l)}$ are typically all equal to 1.
However, they can be set to different values if the observations
are made with different accuracy,
this is particularly convenient for classical systems.
Another use of the weights is the possible generalization from a discrete sum
$\sum_{l=1}^{M}\omega^{(l)}$
to a general measure
$\int d\omega$.\footnote{
\label{normalizingToNumberObservations}
Some authors use $\omega^{(l)} = 1/M$ to normalize the fidelity to the range $[0: 1]$.
However, this approach is inconvenient because $\mathcal{F}$ is no longer an
\href{https://en.wikipedia.org/wiki/Intensive_and_extensive_properties}{extensive quantity}
in the sense that, for two sets of observations, the total fidelity is no longer the sum of the two --- losing its additive property.
For this reason, normalizing to the number of observations is preferred over normalizing to $[0: 1]$.
}

The inverse problem is to find a partially
unitary 
operator $\mathcal{U}$ (a matrix $u_{jk}$ of dimension $D\times n$)
acting from $\mathbf{x}$ to $\mathbf{f}$
\begin{align}
f_j&=\sum\limits_{k=0}^{n-1}u_{jk}x_k
& j=0\dots D-1
\label{fProjXDifferently}
\end{align}
that maximizes the total probability (fidelity)
\begin{align}
\mathcal{F}&
=\sum\limits_{l=1}^{M} \omega^{(l)}
\Big|\Braket{\phi_l|\mathcal{U}|\psi_l}\Big|^2
\xrightarrow[{\mathcal{U}}]{\quad }\max
\label{allProjUKxfAppendix} \\
\Braket{f_j|f_{j^{\prime}}}&= \sum\limits_{k,k^{\prime} =0}^{n-1}u_{jk}
\Braket{x_k|x_{k^{\prime}}}
u^*_{j^{\prime}k^{\prime}}
    \label{optimmatrixConstraintAppendixNUAS}
\end{align}
subject to scalar product preservation (\ref{optimmatrixConstraintAppendixNUAS})
where $j,j^{\prime}=0\dots D-1$.
The operator $\mathcal{U}$ acts from
Hilbert space $\mathbf{x}$ to Hilbert space $\mathbf{f}$,
and it can be viewed as a memoryless
\href{https://en.wikipedia.org/wiki/Quantum_channel}{quantum channel}.
The functional (\ref{allProjUKxfAppendix}) has matrix element
\footnote{
Note that the bra-ket notation in 
$\Braket{\phi|\mathcal{U}|\psi}$ (and (\ref{mlproblemVectorNorm}))
may differ from the measured sample average, such as
(\ref{SfunctionalSummedExpansion})
(which is the sum over $l=1\dots M$).
They are the same in Section \ref{classicMapping}
but different in other examples;
this is an important advance from \cite{malyshkin2019radonnikodym},
where two averages were always considered the same.
For example, if in (\ref{optimmatrixConstraintAppendixNUAS}) the
$\Braket{f_j|f_{j^{\prime}}}$ and $\Braket{x_k|x_{k^{\prime}}}$
are considered as sample averages,
the Gram matrix quantum channel of Section \ref{classicMapping} is obtained.
If they are postulated to be unit matrices,
the traditional unitary learning of Section \ref{traditionalUL} is obtained.
The $S_{jk;j^{\prime}k^{\prime}}$ is always calculated
from  the measured sample as the sum over $l=1\dots M$.
}
$\Braket{\phi_l|\mathcal{U}|\psi_l}$ absolute value squared,
which allows expressing the fidelity
 only with $\Braket{f_j | f_{j^{\prime}}}$, $\Braket{x_k | x_{k^{\prime}}}$
and fourth  order terms $\Braket{f_j x_k | f_{j^{\prime}}x_{k^{\prime}}}$,
without using unavailable ``cross-moments''
$\Braket{x_k|f_j}$.
The functional also exhibits proper wavefunction phase invariance.
A specific aspect of quantum measurement is that the wavefunction
itself cannot be measured; only the squared modulus of the wavefunction
(probability density) can be obtained from a
\href{https://en.wikipedia.org/wiki/Measurement_in_quantum_mechanics}{measurement operation}.
The $\left|\psi\right|^2$ is observable, whereas $\psi$ is not.
Measured wavefunction pairs (\ref{mlproblemVector}) should be considered
as measured $\left|\psi\right|^2$ and $\left|\phi\right|^2$
with the square root applied.
The obtained result is possibly multiplied
by an unknown random phase $\exp(i\xi_l)$.
The optimization problem (\ref{allProjUKxfAppendix}) of finding
the optimal quantum channel $\mathcal{U}$ is invariant with respect to
random phases introduced to measured wavefunctions (\ref{mlproblemVector}).
Without loss of generality, we can consider the Hilbert space bases
to be orthogonal:
\begin{subequations}
\label{ortBasis}
\begin{align}
\delta_{jj^{\prime}}&=\Braket{f_j|f_{j^{\prime}}}\label{ortF}\\
\delta_{kk^{\prime}}&=\Braket{x_k|x_{k^{\prime}}}\label{prtX}
\end{align}
\end{subequations}
if this is not the case --- an orthogonalizing procedure
can be applied, see Appendices A and E of \cite{malyshkin2019radonnikodym}
or one can simply apply
an orthogonalization method of
\href{https://en.wikipedia.org/wiki/Gram%E2%80%93Schmidt_process}{Gram--Schmidt}
type, see
\texttt{\seqsplit{com/polytechnik/algorithms/DemoRecoverMapping.java:GramSchmidtTest}}
for an implementation.
For orthogonal bases,
condition (\ref{optimmatrixConstraintAppendixNUAS})
is exactly the unitary condition if $n=D$. If $D<n$ we name
it the condition of
partial unitarity.
\begin{align}
\delta_{jj^{\prime}}&= \sum\limits_{k=0}^{n-1}u_{jk}
u^*_{j^{\prime} k}& j,j^{\prime}=0\dots D-1
    \label{partialUnitarityOrtBasis}
\end{align}
The meaning of $\mathcal{F}$ (\ref{allProjUKxfAppendix})
is a weighted sum (with $\omega^{(l)}$ weights)
of the possible similarity between $\Ket{\mathcal{U}|\psi_l}$
and $\Ket{\phi_l}$. By construction, we cannot
directly project the states belonging to
different Hilbert spaces as $\Braket{\phi_l|\psi_l}$.
Additionally, such a direct projection will not be invariant with respect to random phases of measured wavefunctions.
The only possible way to transform a state from $\mathbf{x}$ to $\mathbf{f}$
is to apply operator $\mathcal{U}$ (\ref{fProjXDifferently}).
This is a quantum channel that links two different Hilbert spaces.
The states $\psi_l(\mathbf{x})$ and $\phi_l(\mathbf{f})$
in basis (\ref{baseDefXF}) are defined 
with $l=1\dots M$ coefficients $\alpha^{(l)}_k$ and $\beta^{(l)}_j$.
For orthogonalized basis (\ref{ortBasis}) the functional (\ref{allProjUKxfAppendix})
takes the form
\begin{align}
\mathcal{F}&
=\sum\limits_{j,j^{\prime}=0}^{D-1}\sum\limits_{k,k^{\prime}=0}^{n-1}
u_{jk}S_{jk;j^{\prime}k^{\prime}}u^*_{j^{\prime}k^{\prime}}
\label{SfunctionalSummedL}\\
S_{jk;j^{\prime}k^{\prime}}&=
\sum\limits_{l=1}^{M} \omega^{(l)}
\beta^{(l)}_j \alpha^{(l)}_k
\beta^{(l)\, *}_{j^{\prime}} \alpha^{(l)\, *}_{k^{\prime}}
\label{SfunctionalSummedExpansion}
\end{align}
This expression is obtained
by converting $\psi_l(\mathbf{x})$
to a function in $\mathbf{f}$-space
using (\ref{fProjXDifferently})
and then projecting it onto $\phi_l(\mathbf{f})$.
For non-orthogonal bases (\ref{ortBasis}),
the expression for $S_{jk;j^{\prime}k^{\prime}}$
is more complicated and less convenient to use.
The tensor $S_{jk;j^{\prime}k^{\prime}}=S^*_{j^{\prime}k^{\prime};jk}$ is Hermitian
by construction.
The original problem is now reduced to maximizing the functional in
(\ref{SfunctionalSummedL})
subject to the partial unitarity constraint in
(\ref{partialUnitarityOrtBasis}).
There are a number of practical problems that can be reduced to
this optimization problem.

\subsection{\label{quantumTimeEvolution}A Quantum System Time Evolution}

Consider a quantum system with a time-independent
\href{https://en.wikipedia.org/wiki/Hamiltonian_(quantum_mechanics)}{Hamiltonian}
$H$.
Its time evolution
\begin{align}
  \mathcal{U}&=
  \exp \left[-i\frac{t}{\hbar} H \right] \label{Uquantum} \\
   \Ket{\psi^{(t)}}&=\Ket{\mathcal{U} \middle| \psi^{(t=0)}} \label{unitaryPsiEvolution}
\end{align}
Assume we have a long sequence $l=1\dots M$ of system observations
made equidistantly at time moments $t_l=\tau l$
\begin{align}
\psi_l(\mathbf{x})& \to \psi_{l+1}(\mathbf{x})
  & \text{weight $\omega^{(l)}=1$}  \label{mlproblemVectorTIMEEvolution}
\end{align}
Measured wavefunctions can possibly have random phases.
Given the basis (\ref{baseX}), a measured sample
is a sequence of
$l=1\dots M$ coefficients $\alpha^{(l)}_k$ that define the actual wavefunction
within a random phase $\exp(i\xi_l)$.
To obtain the optimization problem of the previous section, we put
 $D=n$, $\beta^{(l)}_j=\alpha^{(l+1)}_k$,
and $\mathcal{U}$ is a unitary operator of wavefunction time shift $\tau$.
\begin{align}
\Ket{\psi_{l+1}}&=\Ket{\mathcal{U}|\psi_l}
\label{UtimeEvolution}
\end{align}
The $S_{jk;j^{\prime}k^{\prime}}$ is then
\begin{align}
S_{jk;j^{\prime}k^{\prime}}&=
\sum\limits_{l=1}^{M} \omega^{(l)}
\alpha^{(l+1)}_j \alpha^{(l)}_k
\alpha^{(l+1)\, *}_{j^{\prime}} \alpha^{(l)\, *}_{k^{\prime}}
\label{SfunctionalSummedExpansionTAUShift}
\end{align}
The result of the optimization problem (\ref{SfunctionalSummedL})
subject to the unitary ($n=D$) constraint (\ref{partialUnitarityOrtBasis})
is a unitary time shift operator $\mathcal{U}$.
To obtain the Hamiltonian from $\mathcal{U}$
requires taking the logarithm of a unitary matrix.
This is a different problem that requires
separate consideration\cite{loring2014computing}.
\begin{align}
H&=i\frac{\hbar}{\tau} \ln \mathcal{U}
\label{logUCalc}
\end{align}

\subsection{\label{classicMapping}A Classical System $\mathbf{x}\to \mathbf{f}$ Vector Mapping}

Consider classical system
$\mathbf{x}^{(l)} \to \mathbf{f}^{(l)}$
vectors mapping with the weights $\omega^{(l)}$, $l=1\dots M$, and all measurements are real numbers.
\begin{align}
  (x_0,x_1,\dots,x_k,\dots,x_{n-1})^{(l)}&\to
  (f_0,f_1,\dots,f_j,\dots,f_{D-1})^{(l)}
   \label{mlproblemVectorObservableValues}
\end{align}
This is a mapping of some observable to observable classical measurements
of real vectors,
for example let $\mathbf{x}$ be attributes
and $\mathbf{f}$ be class labels of some
\href{https://en.wikipedia.org/wiki/List_of_datasets_for_machine-learning_research}{dataset}
used in machine learning \href{https://en.wikipedia.org/wiki/Statistical_classification}{classification problem} of vector type. A few example of  $\mathbf{f}(\mathbf{x})$
constructed
models include 
linear regression,
the Radon-Nikodym approximation\cite{malyshkin2019radonnikodym},
a logical function, a neural network model, etc.

We want to convert these data to 
$\psi_l(\mathbf{x}) \to \phi_l(\mathbf{f})$
from (\ref{mlproblemVector}) to construct a partially
unitary operation $\mathcal{U}$ converting a state from $\mathbf{x}$ to $\mathbf{f}$.
Let us define an average in both Hilbert spaces as
the sum over the data sample (\ref{mlproblemVectorObservableValues})
\begin{align}
  \Braket{h}&=\sum\limits_{l=1}^{M}  \omega^{(l)} h_l 
  \label{gfaverageDef} \\
  G_{kk^{\prime}}^{\mathbf{x}} &= \Braket{x_k | x_{k^{\prime}}} \label{GramX} \\
G_{jj^{\prime}}^{\mathbf{f}} &= \Braket{f_j | f_{j^{\prime}}} \label{GramF}
\end{align}
where $h$ is a function on $\mathbf{x}$ or $\mathbf{f}$.
When (\ref{gfaverageDef}) is applied to $h_l=x^{(l)}_k x^{(l)}_{k^{\prime}}$ or $f^{(l)}_j f^{(l)}_{j^{\prime}}$ 
we obtain the
\href{https://en.wikipedia.org/wiki/Gram_matrix}{Gram matrix}
(\ref{GramX}) or (\ref{GramF}).
For complex space $h_l$ would be $x^{(l)}_k x^{(l)\,*}_{k^{\prime}}$ or $f^{(l)}_j f^{(l)\,*}_{j^{\prime}}$, but in this section $x_k$ and $f_j$ are all real.
As we discussed above, the $\mathbf{x}$ and $\mathbf{f}$
bases can be always
orthogonalized (\ref{ortBasis}) with a basis linear transform;
the Gram matrix is equal to the unit matrix in the case of an orthogonal basis.
Note that with these classical data, we can calculate
``cross-moments'' $\Braket{x_k|f_j}$.
An example of a model that uses cross-moments
is linear regression
\begin{align}
&\min\limits_{\gamma_k}\Braket{\left|f_j-\sum_{k=0}^{n-1}\gamma_k x_k\right|^2}\label{lsqmin}\\
&f_j(\mathbf{x})\approx\sum\limits_{k,k^{\prime}=0}^{n-1} x_kG_{kk^{\prime}}^{\mathbf{x};\,-1} \Braket{f_j | x_{k^{\prime}}}
\label{fxapproxLS}
\end{align}
As we use quantum channel ideology, our model
should not depend on ``cross-moments'',
only the partially unitary operator $\mathcal{U}$ (\ref{fProjXDifferently})
can link $\mathbf{x}$ and $\mathbf{f}$.

To construct a 
wavefunction $\psi_{\mathbf{y}}(\mathbf{x})$ localized at
$\mathbf{x}=\mathbf{y}$ consider
a localized state $\psi_{\mathbf{y}}(\mathbf{x})$
\begin{align}
       \psi_{\mathbf{y}}(\mathbf{x})&
           =\sqrt{K(\mathbf{y})}\sum\limits_{i,k=0}^{n-1}y_iG^{\mathbf{x};\,-1}_{ik}x_k
           \nonumber \\
           &=
           \frac{\sum\limits_{i,k=0}^{n-1}y_iG^{\mathbf{x};\,-1}_{ik}x_k}
           {\sqrt{\sum\limits_{i,k=0}^{n-1}y_iG^{\mathbf{x};\,-1}_{ik}y_k}}             
       =\frac{\sum\limits_{i=0}^{n-1}\psi^{[i]}(\mathbf{y})\psi^{[i]}(\mathbf{x})}
           {\sqrt{\sum\limits_{i=0}^{n-1}\left[\psi^{[i]}(\mathbf{y})\right]^2}}
  \label{psiYlocalized} \\
    K(\mathbf{x})&=\frac{1}{\sum\limits_{i,k=0}^{n-1}x_iG^{\mathbf{x};\,-1}_{ik}x_k}
  =\frac{1}{\sum\limits_{i=0}^{n-1}\left[\psi^{[i]}(\mathbf{x})\right]^2}
  \label{ChristoffelFunctionDef}
\end{align}
The $\psi_{\mathbf{y}}(\mathbf{x})$ is a function on $\mathbf{x}$ localized at
a given $\mathbf{y}$,
it is a normalized
\href{https://en.wikipedia.org/wiki/Reproducing_kernel_Hilbert_space}{reproducing kernel},
 $1=\Braket{\psi_{\mathbf{y}}|\psi_{\mathbf{y}}}$.
In (\ref{psiYlocalized}) it is written in two bases:
the original $x_k$, for which $\Braket{x_qx_k}=G^{\mathbf{x}}_{qk}$,
and in some orthogonalized  basis $\Ket{\psi^{[i]}}$, such that
$\Braket{\psi^{[q]}|\psi^{[k]}}=\delta_{qk}$.
The $K(\mathbf{x})$ is the Christoffel function.
Localized states in $\mathbf{f}$-space can be obtained
with $n$ to $D$ and $x$ to $f$ replacement.
With these localized states, we obtain $\psi_l(\mathbf{x}) \to \phi_l(\mathbf{f})$
mapping for classical data (\ref{mlproblemVectorObservableValues})
with observation weights $\omega^{(l)}$,
$l=1\dots M$
\begin{align}
&\sqrt{K(\mathbf{x}^{(l)})}\sum\limits_{i,k=0}^{n-1}x^{(l)}_iG^{\mathbf{x};\,-1}_{ik}x_k
\to
\sqrt{K(\mathbf{f}^{(l)})}\sum\limits_{i,j=0}^{D-1}f^{(l)}_iG^{\mathbf{f};\,-1}_{ik}f_k
  \label{mlproblemVectorCladssicProblemPsi}
\end{align}
This mapping is actually an original $\mathbf{x}\to\mathbf{f}$
mapping (\ref{mlproblemVectorObservableValues})
with $\mathbf{x}$ and $\mathbf{f}$ linearly transformed and normalized.
Contrary to the original mapping,
(\ref{mlproblemVectorCladssicProblemPsi})
can be considered as a mapping of two Hilbert space states
with the scalar product (\ref{gfaverageDef}).
It is convenient to introduce the moments
of Christoffel function products:
\begin{align}
&\Braket{x_{k}f_{j}|K^{(\mathbf{x})}K^{(\mathbf{f})}|x_{k^{\prime}}f_{j^{\prime}}}=
\nonumber \\
&\sum\limits_{l=0}^{M}\omega^{(l)}
\frac{x^{(l)}_{k}x^{(l)}_{k^{\prime}}}
{
         \sum\limits_{q,q^{\prime}=0}^{n-1}
         x^{(l)}_{q} G^{\mathbf{x};\,-1}_{qq^{\prime}}x^{(l)}_{q^{\prime}}
}
\cdot
\frac{f^{(l)}_{j}f^{(l)}_{j^{\prime}}}
{
\sum\limits_{s,s^{\prime}=0}^{D-1} f^{(l)}_{s}G^{\mathbf{f};\,-1}_{ss^{\prime}}f^{(l)}_{s^{\prime}}
}
\label{ChristoffelfunctionsProductMoments}
\end{align}
Assuming the  $\mathbf{x}$ and $\mathbf{f}$
bases are orthogonalized (\ref{ortBasis}),
the tensor (\ref{SfunctionalSummedExpansion}) is then
\begin{align}
S_{jk;j^{\prime}k^{\prime}}&=\Braket{x_{k}f_{j}|K^{(\mathbf{x})}K^{(\mathbf{f})}|x_{k^{\prime}}f_{j^{\prime}}}
\label{Sclassic}
\end{align}
We have now obtained, for classical data, an optimization problem 
(\ref{SfunctionalSummedL})
subject to the (\ref{partialUnitarityOrtBasis}) partial unitarity
constraint. The result is the operator $\mathcal{U}$ maximizing
$\mathcal{F}$. This problem has the same mathematical form as
the quantum problem of Section \ref{quantumTimeEvolution} above.
But there is one important difference: where the Hilbert space inner product
comes from.
In the quantum problem, it is an original property of Hilbert spaces used to formulate the problem, but the functional
$\mathcal{F}$ is
obtained from measurement data.
In the classical problem, both the scalar product in the Hilbert space and the functional $\mathcal{F}$
are obtained from measurement data.

To evaluate the result at some given point $\mathbf{y}$,
construct the localized density (\ref{psiYlocalized})
and transform it to $\mathbf{f}$-space with (\ref{fProjXDifferently}).
By projecting it to a $\mathbf{g}$-localized state in $\mathbf{f}$-space,
obtain the probability of a given outcome $\mathbf{g}$
(here $\psi_{\mathbf{g}}(\mathbf{f})$ is a $\mathbf{f}$-space state localized
at $\mathbf{f}=\mathbf{g}$, similarly to an $\mathbf{x}$-state (\ref{psiYlocalized}),
i.e. in $g_{i}  G^{\mathbf{f};\,-1}_{ij} f_j$  put $f_j$ from (\ref{fProjXDifferently})
after that obtained expression is a function on $x_k$
that can be coupled with (\ref{psiYlocalized});
for orthogonal bases (\ref{ortBasis})
obtain (\ref{Sclassic}))
\begin{align}
  \Braket{\psi_{\mathbf{g}}|\mathcal{U}|\psi_{\mathbf{y}}}^2
&=
  \frac{
 \left|
    \sum\limits_{k=0}^{n-1}\sum\limits_{j,s=0}^{D-1}
g_{j}  G^{\mathbf{f};\,-1}_{js}
u_{sk}
y_{k}
    \right|^2
  }
       {
         \sum\limits_{j,j^{\prime}=0}^{D-1} g_{j}G^{\mathbf{f};\,-1}_{jj^{\prime}}g_{j^{\prime}}
         \sum\limits_{k,k^{\prime}=0}^{n-1}
         y_{k} G^{\mathbf{x};\,-1}_{kk^{\prime}}y_{k^{\prime}}
       } \label{probgUypkAppAS}        
\end{align}
The optimization result is the $u_{jk}$ matrix, $j=0\dots D-1; k=0\dots n-1$.
This operator, given some input state (such as a localized state $\Ket{\psi_{\mathbf{x}}}$),
uniquely (within a phase) finds the function
in $\mathbf{f}$-space
$\Ket{\mathcal{U}|\psi_{\mathbf{x}}}$
(coefficients $a_j(\mathbf{x})$)
that predicts
the probability (\ref{probgUypkAppAS})
of outcome $\mathbf{f}$:
\begin{align}
P(\mathbf{f})\Big|_{\mathbf{x}}&=
\Braket{\psi_{\mathbf{f}}|\mathcal{U}|\psi_{\mathbf{x}}}^2
= \frac{\left|\sum\limits_{j=0}^{D-1} a_jf_j\right|^2}
{
\sum\limits_{j,j^{\prime}=0}^{D-1}f_jG^{\mathbf{f};\,-1}_{jj^{\prime}}f_{j^{\prime}}
} \label{probFXUpExpanded} \\
a_j&=
\sum\limits_{s=0}^{D-1}
\sum\limits_{k=0}^{n-1} G^{\mathbf{f};\,-1}_{js} u_{sk}x_k\sqrt{K(\mathbf{x})}
\label{aOnXCoefs}
\end{align}
the $\mathbf{f}$ is
equal to the value of the outcome
we are interested in determining the probability of.
Given $\mathbf{x}$, the probability of an outcome  $\mathbf{f}$
is a squared linear function on $f_j$ multiplied by the Christoffel function
$K(\mathbf{f}$).
This form of probability,
a squared linear function on $f_j$ divided by a quadratic form on $f_j$,
can be obtained from many different considerations;
the difference between them lies in the coefficients $a_j$.
The mapping (\ref{mlproblemVectorCladssicProblemPsi}) corresponds to pure states.
For mixed states mapping, the probability will be
a ratio of two general quadratic forms instead of a rank one matrix
in the numerator (\ref{probFXUpExpanded}).

For a simple demonstration of creating 
$\mathbf{x}\rightarrow\mathbf{f}$ classical mapping,
refer to Section \ref{funDemo} below.

\subsection{\label{UnitaryLearning}Learning Unitary Dynamics}
In nature most of the dynamic equations
are equivalent to a sequence of infinitesimal unitary transformations:
Newton, Maxwell, and Schr\"{o}dinger equations.
 Consider a simple
classical problem.
Let there be an initial state vector 
$\mathbf{X}^{(0)}$
of unit $L^2$ norm
and a unitary matrix $\mathcal{U}$.
The operator $\mathcal{U}$ is applied to $\mathbf{X}^{(0)}$
$M$ times:
\begin{align}
\mathbf{X}^{(l+1)}&=\mathcal{U}\mathbf{X}^{(l)}
\label{unitaryDynamcs}
\end{align}
From this sequence, $M$ observations $(\mathbf{x},\mathbf{f})$
(\ref{mlproblemVectorObservableValues}) are created by
taking $(l,l+1)$ elements of the sequence and multiplying them
by a random phase (or $\pm 1$ for real space).
\begin{subequations}
\label{timeEvolutionSample}
\begin{align}
\mathbf{x}^{(l)}&=\exp(i\xi_l) \mathbf{X}^{(l)} \label{xfromX}\\
\mathbf{f}^{(l)}&=\exp(i\zeta_l)\mathbf{X}^{(l+1)} \label{ffromX}
\end{align}
\end{subequations}
The random phases make any $\mathbf{x}\leftrightarrow\mathbf{f}$
regression-type methods inapplicable.\footnote{
\label{fnRandomPhase}
Consider the data in (\ref{timeEvolutionSample}). Then
 $\Braket{f_j | x_k}=
\sum\limits_{l=1}^{M} \omega^{(l)}
f^{(l)}_j x^{(l)\,*}_k \exp(i(\zeta_l-\xi_l))$
which depends on random phases $\zeta_l$ and $\xi_l$, and thus it cannot be used to reconstruct the system.
}
From the unitary property of operator $\mathcal{U}$,
we immediately obtain (\ref{optimmatrixConstraintAppendixNUAS});
it does not depend on random phases.
Whereas in the quantum problem we have a Hilbert space with
an inner product $\Braket{\cdot}$,
in this classical problem there is no built in inner product available.
The only available average is the sum (\ref{gfaverageDef})
over $M$ observations;
there is no any other average such
as ensemble averages or quantum measurements.
For unitary dynamic with $\omega^{(l)}=1$,
this sum represents the regular time-average.
The problem is to determine the operator  $\mathcal{U}$
from the sample (\ref{timeEvolutionSample}), $l=0\dots M-1$,
that undergoes unitary time-evolution (\ref{unitaryDynamcs}).
The goal is to determine a matrix
 $u_{jk}$ (with $D=n$) maximizing the quality criterion $\mathcal{F}$
(\ref{allProjUKxfAppendix}).

The Gram matrices $G_{kk^{\prime}}^{\mathbf{x}}$ (\ref{GramX}) and $G_{jj^{\prime}}^{\mathbf{f}}$ (\ref{GramF})
are time-average.
The functional $\mathcal{F}$ 
is also time-average of the (\ref{unitaryDynamcs}) data
\begin{align}
\mathcal{F}&
=\sum\limits_{l=1}^{M} \omega^{(l)}
\left|
\sum\limits_{j=0}^{D-1}\sum\limits_{k=0}^{n-1}
X^{(l+1)}_j u_{jk}X^{(l)}_k
\right|^2
\xrightarrow[{u_{jk}}]{\quad }\max
\label{classicUnitaryDynamics} 
\end{align}
Using (\ref{timeEvolutionSample}) obtain
\begin{align}
S_{jk;j^{\prime}k^{\prime}}&=
\sum\limits_{l=1}^{M} \omega^{(l)}
f^{(l)}_j x^{(l)}_k
f^{(l)\,*}_{j^{\prime}} x^{(l)\,*}_{k^{\prime}}
\label{SfunctionalClassic}
\end{align}
Random phases ($\pm 1$ for real space)
in (\ref{timeEvolutionSample})  do not affect $S_{jk;j^{\prime}k^{\prime}}$
and the Gram matrices
 $\Braket{x_k | x_{k^{\prime}}}$,
$\Braket{f_j | f_{j^{\prime}}}$
as the phases cancel each other in probabilities.
The Gram matrices are not necessary unit matrices.
This can be changed by basis regularization.
Introduce regularization transformations
$R^{\mathbf{x}}$
and
$R^{\mathbf{f}}$
such that
\begin{subequations}
\label{regularization}
\begin{align}
\mathfrak{x}&=R^{\mathbf{x}}{\mathbf{x}} \label{regularizationX}\\
\mathfrak{f}&=R^{\mathbf{f}}{\mathbf{f}} \label{regularizationF}
\end{align}
\end{subequations}
produce unit Gram matrices;
this can be for example $R^{\mathbf{x}}=G^{\mathbf{x}; -1/2}$,
$R^{\mathbf{f}}=G^{\mathbf{f}; -1/2}$ or any other method,
such as Gram-Schmidt with pivoting
or \href{https://en.wikipedia.org/wiki/QR_decomposition}{QR decomposition}.
Then transformation (\ref{fProjXDifferently})
can be written in the form
\begin{align}
\mathfrak{f}&=
R^{\mathbf{f}}\mathcal{U} R^{\mathbf{x};-1} \mathfrak{x}
\label{nBasis}
\end{align}
and we can consider an optimization problem
for the operator $\widetilde{\mathcal{U}}=R^{\mathbf{f}}\mathcal{U} R^{\mathbf{x};-1}$
instead of for $\mathcal{U}$. The solution in the original basis
is then $\mathcal{U}=R^{\mathbf{f};-1}\widetilde{\mathcal{U}} R^{\mathbf{x}}$,
see
\texttt{\seqsplit{com/polytechnik/algorithms/DemoRecoverUnitaryMatrixFromSeq.java:getUFromConfigGramMatrixChannel}} for an implementation.

A question can be asked: In the case where $D=n$ 
for the data (\ref{timeEvolutionSample}),
will an operator of system dynamics
$\mathcal{U}$ having
the matrix $u_{jk}$ maximizing $\mathcal{F}$
subject to constraints (\ref{optimmatrixConstraintAppendixNUAS})
always be unitary?
 It depends on the data.
For a dataset representing unitary dynamics (\ref{unitaryDynamcs})
the Gram matrices  $\Braket{x_k | x_{k^{\prime}}}$
and
$\Braket{f_j | f_{j^{\prime}}}$
must be the same, since the unitary operator
$\mathcal{U}$ preserves the scalar product.
But if the data contain non-unitary contributions,
the Gram matrices
 $\Braket{x_k | x_{k^{\prime}}}$,
$\Braket{f_j | f_{j^{\prime}}}$
can be different, and this difference contributes to
the non-unitarity of $u_{jk}$;
the constraints (\ref{optimmatrixConstraintAppendixNUAS})
preserve the Gram matrix passing through the quantum channel.
This is the meaning
of the constraints: the Gram matrix
must transform from the Hilbert space 
\emph{IN} 
into
the Hilbert space 
\emph{OUT}
like any other operator.
The idea is to measure the Gram matrix from the sample in both Hilbert spaces,
construct $u_{jk}$ by solving the optimization problem, and then
generalize the model, stating that any other operator
converts in the same way (\ref{operatorTransform}).
If operators other than the Gram matrix are available in both Hilbert spaces, they can be used to build a quantum channel in exactly the same manner.
Consider the unit matrix
transformation, which corresponds to traditional unitary learning.

\subsection{\label{traditionalUL} Traditional Unitary Learning}
Consider ``traditional'' unitary learning.
For the data sample (\ref{timeEvolutionSample}),
it is postulated that the operator $\mathcal{U}$
(\ref{unitaryDynamcs}) is unitary;
hence the Gram matrix properties
are irrelevant to the task.
All observation vectors $\mathbf{x}$ and $\mathbf{f}$
are of unit $L^2$ norm and of the same dimension.
The problem becomes the following:
maximize $\mathcal{F}$ (\ref{SfunctionalSummedL})
subject to unitary $\mathcal{U}$, i.e. 
constraints (\ref{partialUnitarityOrtBasis})
with $D=n$.
This is exactly the problem considered
in Section \ref{UnitaryLearning} above,
but with a quantum channel transforming
(\ref{operatorTransform})
a unit matrix from Hilbert space 
\emph{IN} 
into a unit matrix
in Hilbert space \emph{OUT}
instead of a Gram matrix.
Practically, this unit matrix quantum channel can be implemented
exactly as considered above. The only difference is that
no regularization (\ref{regularization})
should be performed at all, since it is postulated that
$\mathcal{U}$ must always be unitary.
The $\mathbf{f}$ and $\mathbf{x}$ should be used directly as is
(without regularization)
when constructing $S_{jk;j^{\prime}k^{\prime}}$ (\ref{SfunctionalClassic}),
see
\texttt{\seqsplit{com/polytechnik/algorithms/DemoRecoverUnitaryMatrixFromSeq.java:getUFromConfigUnitMatrixChannel}} for an implementation.
Contrary to the Gram matrix quantum channel of the previous section,
the $\mathcal{U}$ obtained with the unit matrix quantum channel
is always unitary.
For a simple demonstration of recovering $u_{jk}$ from unitary dynamics data
(\ref{timeEvolutionSample}) refer to
Section \ref{recoverUnitaryMatrix} below.

This unitary learning considers
$\mathbf{x}$ and $\mathbf{f}$ vectors to be of \textsl{the same} dimension.
The approach can be directly generalized to partial unitarity.
Assume a dataset is of  $\mathbf{x}\to\mathbf{f}$
mapping,
where all vectors 
 $\mathbf{x}$ and $\mathbf{f}$
have unit $L^2$ norm, but the vectors $\mathbf{x}$ and $\mathbf{f}$
are  of \textsl{different} dimensions:
$n$ and $D$ respectively.
We want to build a \textsl{partially unitary} operator $\mathcal{U}$
of dimension $D\times n$ that converts a vector
from $\mathbf{x}$ to $\mathbf{f}$ preserving probability.
The model makes direct assumption about $\mathcal{U}$ and the dataset,
hence Gram matrix properties are irrelevant to the task.
The problem becomes the following:
maximize $\mathcal{F}$ (\ref{SfunctionalSummedL})
subject to partial unitary constraints (\ref{partialUnitarityOrtBasis}),
 now with $D<n$. The calculations are identical to the problem where $D=n$
that we just considered.
Calculate $S_{jk;j^{\prime}k^{\prime}}$ (\ref{SfunctionalClassic})
using $\mathbf{x}$ and $\mathbf{f}$ directly, without regularization.
Then optimize $\mathcal{F}$ subject to (\ref{partialUnitarityOrtBasis})
with corresponding $D$ and $n$.
The obtained partially unitary operator
$\mathcal{U}$
is a quantum channel transforming a unit matrix
of dimension $n$ in Hilbert space $\mathbf{x}$
to a unit matrix of dimension $D$ in Hilbert space $\mathbf{f}$.
For a simple demonstration, refer to Section \ref{funDemoNcNx} below.
This quantum channel
maximizes the fidelity
of mapping between Hilbert spaces of different dimensions.
This can offer a completely new perspective on unitary machine learning models.

\subsection{\label{OOptU}Variational quantum algorithms}
Variational quantum algorithms\cite{cerezo2021variational}, which use a classical optimizer to train a parametrized quantum circuit, are among the most promising applications of near-term quantum computers.
They often have a cost function in the form\cite{park2024hamiltonian}
\begin{align}
C(\theta)&=
\mathrm{Tr}\, O \mathcal{U}(\theta)\rho_0\mathcal{U}^{\dagger}(\theta)
\label{costVQA}
\end{align}
where $O$ is a Hermitian operator.
This expression is a quadratic function of the unitary operator $\mathcal{U}$.
Expanding the trace, we obtain the cost function exactly in the form (\ref{allProjUKxfDIAG}),
subject to unitary constraints (\ref{optimmatrixConstraintAppendixNUDIAG}).
The tensor $S_{jk;j^{\prime}k^{\prime}}$ is obtained directly from (\ref{costVQA}).
Thus, the theory developed in this paper is directly applicable to variational quantum algorithms.
The application of the developed algorithm is demonstrated for problems of dimension as large as $40$;
see Section \ref{recoverUnitaryMatrix} below.
The limitation now is not due to vanishing gradients or a too-flat cost function\cite{park2024hamiltonian},
but to computational complexity. Our optimization algorithm can be parallelized (see Appendix \ref{compComplexityAnalysis} below),
and we see no significant difficulty preventing an increase in problem dimension.

\subsection{\label{algebraicStructure}Algebraic Structure of the Optimization Problem}

The formulated optimization problem
maximizes (\ref{SfunctionalSummedL}) subject to the partial
unitarity constraint (\ref{partialUnitarityOrtBasis}).
This is a variant of the \href{https://en.wikipedia.org/wiki/Quadratically_constrained_quadratic_program}{QCQP} problem.
We find an operator $\mathcal{U}$
that optimally transforms a state $\Ket{\psi({\mathbf{x}})}$
from Hilbert space \emph{IN} (of dimension $n$)
into a state $\Ket{\phi({\mathbf{f}})}$
from Hilbert space \emph{OUT} (of dimension $D$).
The operator is a rectangular matrix of dimension
$D \times n$ that transforms an operator $A$
from \emph{IN} to \emph{OUT} as
\begin{align}
A^{\emph{OUT}}&=\mathcal{U} A^{\emph{IN}} \mathcal{U}^{\dagger}
\label{operatorTransform}
\end{align}
This transformation converts any operator between two Hilbert spaces,
for example
a pure state $A^{\emph{IN}}=\Ket{\psi}\Bra{\psi}$
into a pure state $A^{\emph{OUT}}=\Ket{\mathcal{U}|\psi}\Bra{\psi|\mathcal{U}^{\dagger}}$.
For $D=n$ it is a
\href{https://en.wikipedia.org/wiki/Quantum_channel}{trace preserving map}
and for $D<n$ it is a 
\href{https://en.wikipedia.org/wiki/Quantum_channel}{trace decreasing map}
quantum channel (see Fig. \ref{FIgFidelityMax} for a demonstration);
for a more
\href{https://www.youtube.com/watch?v=cMl-xIDSmXI}{general form},
refer to
\href{https://learning.quantum.ibm.com/course/general-formulation-of-quantum-information/quantum-channels#kraus-representations}{Kraus operators} (\ref{KrausOperator}) below.
There should always be an operator known in both Hilbert spaces, which is used to create constraints on $\mathcal{U}$
when maximizing fidelity.
These constraints (probability preservation) determine
the specific form of partial unitarity.
We consider two such operators: the Gram matrix (used throughout most of this paper) and the unit matrix (traditional in unitary learning,
as in Section \ref{traditionalUL},
a \href{https://en.wikipedia.org/wiki/Unital_map}{unital map}).
The constrained optimization problem on $\mathcal{U}$
is reduced to a new algebraic problem (a variation of the Lagrangian 
$\mathcal{L}$
(\ref{lagrangetovariateNUDlen})
is set to zero (\ref{variationUIngivenState}))
\begin{align}
  S \mathcal{U} &= \lambda \mathcal{U}
  \label{eigenvaluesLikeProblem}
\end{align}
which determines the operator $\mathcal{U}$.
It is remotely similar to
the stationary Schr\"{o}dinger equation (eigenvalue problem),
but instead of a Hamiltonian there is a superoperator $S$
(represented by the tensor $S_{jk;j^{\prime}k^{\prime}}$).
The ``eigenvector'' $\mathcal{U}$ is a partially unitary operator
(represented by a $D\times n$ matrix $u_{jk}$),
and the ``eigenvalue'' $\lambda$ is a Hermitian matrix
(represented by a $D\times D$ matrix $\lambda_{ij}$).
The extremal functional value  $\mathcal{F}$ is equal
to the
\href{https://en.wikipedia.org/wiki/Trace_(linear_algebra)}{trace}
(the sum of diagonal elements)
of $\lambda$.
The algebraic structure of this
eigenvalue--like problem,
let us call it an ``\textbf{eigenoperator}'' problem,
requires a separate study.
Currently, we only have a numeric solution algorithm.

\section{\label{numericalSolutionSect}Numerical Solution}

The problem is to optimize (\ref{allProjUKxfDIAG})
subject to constraints (\ref{optimmatrixConstraintAppendixNUDIAG}) 
\begin{align}
\mathcal{F}&=
\sum\limits_{j,j^{\prime}=0}^{D-1}\sum\limits_{k,k^{\prime}=0}^{n-1}
u_{jk}S_{jk;j^{\prime}k^{\prime}}u^*_{j^{\prime}k^{\prime}}
\xrightarrow[{u}]{\quad }\max
\label{allProjUKxfDIAG} \\
\delta_{jj^{\prime}}&= \sum\limits_{k=0}^{n-1}u_{jk}u^*_{j^{\prime}k} \qquad\qquad  j,j^{\prime}=0\dots D-1
    \label{optimmatrixConstraintAppendixNUDIAG}
\end{align}
The tensor $S_{jk;j^{\prime}k^{\prime}}=S^*_{j^{\prime}k^{\prime};jk}$ is Hermitian.
For simplicity, we consider $S_{jk;j^{\prime}k^{\prime}}$ and $u_{jk}$
to be real,
and do not explicitly denote complex conjugation ($^*$)  below,
generalization to complex values is straightforward.
The matrix $u_{jk}$ is an \href{https://en.wikipedia.org/wiki/Isometry}{isometry},
which means it has orthonormal rows but may not be a square matrix.
If we consider a subset of constraints (\ref{optimmatrixConstraintAppendixNUDIAG}),
the optimization problem can be readily solved.
Consider
the squared
\href{https://en.wikipedia.org/wiki/Matrix_norm#Frobenius_norm}{Frobenius norm}
of matrix $u_{jk}$
to be a ``simplified constraint'':
\begin{align}
    &\sum\limits_{j=0}^{D-1}\sum\limits_{k=0}^{n-1}u^2_{jk} =D
    \label{optimmatrixConstraintScalarAppendix}
\end{align}
This is a partial constraint (it is the sum of all diagonal elements in (\ref{optimmatrixConstraintAppendixNUDIAG})).
For this partial constraint, the optimization problem
(\ref{allProjUKxfDIAG})
 is equivalent to an eigenvalue problem ---
it can be directly solved
by considering a vector of $Dn$ dimension
obtained from the $u_{jk}$,
saving all its components into a
\href{https://en.wikipedia.org/wiki/Vectorization_(mathematics)}{single vector},
row by row.

The main idea of \cite{malyshkin2022machine}
was to modify the partial constraint
solution to satisfy the full set of constraints
(\ref{optimmatrixConstraintAppendixNUDIAG}).
There are several options to adjust the solution
from a partial to the full set of constraints.
The one
producing the minimal change to the solution
is the application of the Gram matrix
\begin{align}
G^{u}_{jj^{\prime}}&=\sum\limits_{k=0}^{n-1}u_{jk}u_{j^{\prime}k}
\label{GramUpartial}
\end{align}
inverse square root $G^{u;-1/2}=1/\sqrt{G^{u}}$ to $u_{jk}$.\footnote{
See Appendix A of \cite{malyshkin2022machine}
for adjustments in different bases
and for an approach that uses
\href{https://en.wikipedia.org/wiki/Singular_value_decomposition}{SVD}
instead of the eigenproblem (\ref{GramMatrixEV}).
Also note that both SVD-based and eigenproblem-based
transformations convert a single
 $u_{jk}$ state
satisfying the partial constraint (\ref{optimmatrixConstraintScalarAppendix})
to a state satisfying the
full set of constraints (\ref{optimmatrixConstraintAppendixNUDIAG}).
The problem of converting \textsl{several} $u^{[s]}_{jk}$
states satisfying a partial constraint to a single state satisfying the full set of constraints can greatly improve the initial convergence of the algorithm. This is a subject of future research.
}
There are $2^{D}$ square roots of a
\href{https://en.wikipedia.org/wiki/Definite_matrix}{positively definite}
symmetric matrix of dimension $D$,
differing only in the $\pm$ signs. The simplest method to calculate it
is to
convert $G^{u}_{jj^{\prime}}$ to diagonal form in the basis of its
eigenvectors
\begin{align}
\Ket{G^{u}\middle|g^{[i]}}&=\lambda_G^{[i]}\Ket{g^{[i]}}
\label{GramMatrixEV}
\end{align}
then change the eigenvalues to
$\pm 1\Big/\sqrt{\lambda_G^{[i]}}$
and convert the matrix back to the initial basis
\begin{align}
\left\|G^{u;-1/2}\right\|&=\sum\limits_{i=0}^{D-1}\frac{\pm 1}{\sqrt{\lambda_G^{[i]}}}
\Ket{g^{[i]}}\Bra{g^{[i]}}
\label{mSqrtG}
\end{align}
By checking the result, one can verify that for any $u_{jk}$
producing a non-degenerated Gram matrix (\ref{GramUpartial}), the vector
\begin{align}
\widetilde{u}_{jk}&=
\sum\limits_{i=0}^{D-1}G^{u;-1/2}_{ji}u_{ik}
\label{uAdjEx}
\end{align}
satisfies all the constraints 
(\ref{optimmatrixConstraintAppendixNUDIAG})
(the most general form of adjustment
is an application of $U G^{u;-1/2}$ to $u_{ik}$, where $U$ is an arbitrary
unitary operator; the $\pm$ square root signs
can be included in $U$; below
we consider all signs in (\ref{mSqrtG}) to be ``$+$''),
see \texttt{\seqsplit{com/polytechnik/kgo/AdjustedStateToUnitaryWithEV.java}} for an implementation.
Thus, the multiple constraints optimization  problem (\ref{allProjUKxfDIAG})
can be reduced to an unconstrained optimization problem
(we use the identity $D=\sum_{j=0}^{D-1}\sum_{k=0}^{n-1}u^2_{jk}
=\sum_{j,j^{\prime}=0}^{D-1}\sum_{k=0}^{n-1}u_{jk}G^{u;-1}_{jj^{\prime}}u_{j^{\prime}k}$):
\begin{align}
\mathcal{F}&=\frac{
\sum\limits_{j,j^{\prime},i,i^{\prime}=0}^{D-1}\sum\limits_{k,k^{\prime}=0}^{n-1}
u_{jk}
G^{u;-1/2}_{ji}
S_{ik;i^{\prime}k^{\prime}}
G^{u;-1/2}_{j^{\prime}i^{\prime}}
u_{j^{\prime}k^{\prime}}}
{
\frac{1}{D}\sum\limits_{j=0}^{D-1}\sum\limits_{k=0}^{n-1}u^2_{jk}
}
\xrightarrow[{u}]{\quad }\max
\label{allProjUKxfDIAGGm05}
\end{align}
However, this unconstrained problem\footnote{
There are alternative ways
to formulate
an unconstrained optimization
problem. In the unitary case where $D=n$, one can use
$D(D+1)/2$ parameters to parametrize a Hermitian matrix.
A unitary matrix is then obtained through matrix exponentiation (\ref{Uquantum});
functional optimization, however, requires
the derivatives of the matrix exponent,
which complicates the problem\cite{najfeld1995derivatives,RazaDifferentiation}.
There is an option to parametrize a unitary
matrix using the
\href{https://en.wikipedia.org/wiki/Cayley_transform\#Operator_map}{Cayley transform}
$U= (I-A)(I+A)^{-1}$ for
a skew-Hermitian matrix $A^{\dagger}=-A$,
see for example \cite{lu2023efficient,schafers2024modified}.
Both methods are problematic to use,
especially in the case of rectangular $u_{jk}$ with $D<n$.
See also \cite{RazaUnitaryParametrization,wang2024variational} for other methods
of unitary parametrization.
}
\begin{itemize}
\item
cannot be reduced to an eigenvalue
problem since $G^{u;-1/2}_{ji}$ itself depends on $u_{jk}$.
\item is degenerate: there are multiple $u_{jk}$
producing the same $\mathcal{F}$.
Convert a
solution $u_{jk}$
to the Gram matrix
basis (\ref{GramMatrixEV}), change the eigenvalues, then convert
it back to the initial basis, this is similar
to the transformation in
(\ref{mSqrtG}).
Hence the
\href{https://en.wikipedia.org/wiki/Hessian_matrix}{Hessian matrix}
is degenerated,
which
prevents us from directly applying
\href{https://en.wikipedia.org/wiki/Newton%27s_method_in_optimization}{Newton}
and
\href{https://en.wikipedia.org/wiki/Quasi-Newton_method}{quasi-Newton}
optimization methods.
One can possibly use a penalty function like
$\sum 1/\lambda_G$
\begin{align}
\mathrm{Tr} G^{u;-1}&=\sum_{i=0}^{D-1}\frac{1}{\lambda^{[i]}_G}
\label{penaltyFun}
\end{align}
which has the minimum value $D$  when
all the constraints (\ref{optimmatrixConstraintAppendixNUDIAG}) are satisfied.
\end{itemize}

Alternatively, an iterative approach with Lagrange multipliers can be used.
Iterations are required since we cannot simultaneously
solve the equation for $u_{jk}$ and $\lambda_{jj^{\prime}}$.
A single iteration consists in
solving an eigenvalue problem, adjusting
the obtained solution to satisfy the full set of constraints,
and calculating the new values of Lagrange multipliers.
There are three main elements of the algorithm:
\begin{itemize}
\item The eigenproblem solution (\ref{EPLeV}) is used to solve the partially constrained optimization problem.
An important feature
is that any additional linear constraints on $u_{jk}$ of the form
(\ref{linearConstraintsHomog})
can be easily incorporated --- obtain the eigenproblem (\ref{EPLeVV}).
\item The solution adjustment operation (\ref{uAdjEx})
converts a solution satisfying the partial constraint
(\ref{optimmatrixConstraintScalarAppendix})
into one satisfying the full set
 (\ref{optimmatrixConstraintAppendixNUDIAG}).
\item The linear system solution (\ref{newLambdaSolPartial})
obtains the new values of Lagrange multipliers.
\end{itemize}
In its na\"{\i}ve form (without additional linear constraints)
a convergence of the iterative algorithm
turned out to be poor.
The major new result of the current paper is
this iterative algorithm with \textsl{good convergence}.
Good convergence was achieved by considering additional
linear constraints in the eigenproblem at each iteration step.
Consider Lagrange multipliers $\lambda_{jj^{\prime}}$, a matrix
of dimension $D\times D$,
to approach optimization problem (\ref{allProjUKxfDIAG})
with constraints (\ref{optimmatrixConstraintAppendixNUDIAG})
\begin{align}
  \mathcal{L}&=
  \sum\limits_{j,j^{\prime}=0}^{D-1}\sum\limits_{k,k^{\prime}=0}^{n-1}
             u_{jk}S_{jk;j^{\prime}k^{\prime}}u_{j^{\prime}k^{\prime}} \nonumber \\
             &+
   \sum\limits_{j,j^{\prime}=0}^{D-1}        
   \lambda_{jj^{\prime}}\left[\delta_{jj^{\prime}}-\sum\limits_{k^{\prime}=0}^{n-1}u_{jk^{\prime}} u_{j^{\prime}k^{\prime}} \right]
   \xrightarrow[u]{\quad }\max
   \label{lagrangetovariateNUDlen}
\end{align}
Variating $\mathcal{L}$ over $u_{sq}$ obtain (\ref{variationU})
\begin{align}
b_{sq}&=\sum\limits_{j^{\prime}=0}^{D-1}\sum\limits_{k^{\prime}=0}^{n-1}
S_{sq;j^{\prime}k^{\prime}}u_{j^{\prime}k^{\prime}}
\label{Bsq} \\
 0&= 
 b_{sq}-
\sum\limits_{j^{\prime}=0}^{D-1} \lambda_{sj^{\prime}} u_{j^{\prime}q}
\label{variationU}
\end{align}
Here and below we consider the Lagrange multipliers matrix
to be Hermitian
$\lambda_{jj^{\prime}}=\lambda^{*}_{j^{\prime}j}$.
This condition ensures the existence
of an extremal solution\cite{malyshkin2022machine}.
The variation (\ref{variationU}) contains $Dn$
equations. The Hermitian Lagrange multipliers matrix $\lambda_{jj^{\prime}}$
has $D^2$ real parameters ($D(D+1)/2$ independent ones)
for real space and $2D^2$ real parameters ($D^2$ independent ones)
for complex space. Thus, for a general $u_{jk}$,
the variation (\ref{variationU}) cannot be fully satisfied.
The most straightforward way to obtain Lagrange multipliers
for a given $u_{jk}$
is to take the $L^2$
norm of the variation (\ref{variationU}) and obtain the
$\lambda_{ij}$ that minimizes the sum of squares.
\begin{align}
&\sum\limits_{i=0}^{D-1}\sum\limits_{q=0}^{n-1}
\left| b_{iq}-
\sum\limits_{j=0}^{D-1} \frac{\lambda_{ij}+\lambda_{ji}}{2} u_{jq}
\right|^2
 \xrightarrow[\lambda_{ij}]{\quad }\min
\label{lamMinSQuares} \\
&\lambda_{ij}=\mathrm{Herm}
    \sum\limits_{k=0}^{n-1}u_{ik}b_{jk}=
    \mathrm{Herm}
    \sum\limits_{j^{\prime}=0}^{D-1}\sum\limits_{k,k^{\prime}=0}^{n-1}u_{ik}S_{jk;j^{\prime}k^{\prime}}u_{j^{\prime}k^{\prime}}
    \label{newLambdaSolPartial}
\end{align}
The minimization yields a Hermitian matrix $\lambda_{ij}$,
which is obtained
as a solution to a linear system,
see  \texttt{\seqsplit{com/polytechnik/kgo/LagrangeMultipliersPartialSubspace.java:calculateRegularLambda}} for an implementation.
A more general form of $\lambda_{ij}$
is presented in Appendix \ref{LambdaSubspace} below.
The problem can now be considered as maximizing a quadratic form
with the matrix $\mathcal{S}_{jk;j^{\prime}k^{\prime}}$
\begin{align}
  &\mathcal{S}_{jk;j^{\prime}k^{\prime}}=
             S_{jk;j^{\prime}k^{\prime}}
             -\lambda_{jj^{\prime}}\delta_{kk^{\prime}}
   \label{lagrangetovariateNUDlenMatrMax}\\
&\sum\limits_{j,j^{\prime}=0}^{D-1}\sum\limits_{k,k^{\prime}=0}^{n-1}
             u_{jk}\mathcal{S}_{jk;j^{\prime}k^{\prime}}u_{j^{\prime}k^{\prime}}
   \xrightarrow[u]{\quad }\max
   \label{lagrangetovariateNUDlenMax}
\end{align}
subject to constraints (\ref{optimmatrixConstraintAppendixNUDIAG}).
Consider the eigenproblem\footnote{
From the variation (\ref{variationU}), it follows that
the most interesting $u^{[s]}_{jk}$ states have a value of
the functional
(\ref{EPL}) close to zero.
Therefore, 
the matrix $Q_{jj^{\prime}}$ in the denominator can be chosen
as any positively definite Hermitian matrix to improve convergence.
Choosing $Q_{jj^{\prime}}=\delta_{jj^{\prime}}$
gives familiar results; another choice can be
$Q_{jj^{\prime}}=\lambda_{jj^{\prime}}$,
i.e., having the Lagrange multipliers
in the denominator instead of adding them to 
 (\ref{lagrangetovariateNUDlenMatrMax}):
consider
a variation of 
$\sum\limits_{j,j^{\prime}=0}^{D-1}\sum\limits_{k,k^{\prime}=0}^{n-1}
u_{jk}S_{jk;j^{\prime}k^{\prime}}u_{j^{\prime}k^{\prime}}
\Big/
\sum\limits_{j,j^{\prime}=0}^{D-1}\sum\limits_{k=0}^{n-1}
u_{jk}\lambda_{jj^{\prime}}u_{j^{\prime}k}$
over $u_{jk}$ to obtain the same result as (\ref{variationU});
 see \texttt{\seqsplit{com/polytechnik/kgo/KGOIterationalLagrangeMultipliersInDenominatorU.java}}.
}
\begin{align}
&\frac{
\sum\limits_{j,j^{\prime}=0}^{D-1}\sum\limits_{k,k^{\prime}=0}^{n-1}
u_{jk}
\mathcal{S}_{jk;j^{\prime}k^{\prime}}
u_{j^{\prime}k^{\prime}}
}{
\sum\limits_{j,j^{\prime}=0}^{D-1}\sum\limits_{k=0}^{n-1}
u_{jk}Q_{jj^{\prime}}u_{j^{\prime}k}
} 
 \xrightarrow[u]{\quad }\max
 \label{EPL} \\
&   \sum\limits_{j^{\prime}=0}^{D-1}\sum\limits_{k^{\prime}=0}^{n-1}
S_{sq;j^{\prime}k^{\prime}}u^{[s]}_{j^{\prime}k^{\prime}}
-\sum\limits_{j^{\prime}=0}^{D-1} \lambda_{sj^{\prime}} u^{[s]}_{j^{\prime}q}
=\mu^{[s]}\sum\limits_{j^{\prime}=0}^{D-1} Q_{sj^{\prime}} u^{[s]}_{j^{\prime}q}
\label{EPLeV}
\end{align}
with an additional $N_d$ linear constraints added
(their specific form, which provides good convergence of the algorithm, is discussed below in Appendix \ref{linConst})
\begin{align}
0&=\sum\limits_{j=0}^{D-1}\sum\limits_{k=0}^{n-1} 
C_{d;jk}u_{jk}
 & d=0\dots N_d-1
\label{linearConstraintsHomog}
\end{align}
A common method of solving eigenproblem (\ref{EPL})
with
\href{https://en.wikipedia.org/wiki/System_of_linear_equations#Homogeneous_systems}{homogeneous} linear constraints (\ref{linearConstraintsHomog})
is the Lagrange multipliers method\cite{golub1973some}.
This approach, however, creates difficulties when
both linear and quadratic constraints are present,
especially when the number of constraints is large.
A better approach to dealing with linear constraints is to convert
(\ref{linearConstraintsHomog})
to a form that expresses $\mathrm{rank}(C_{d;jk})$ components of $u_{jk}$
in terms of its other components, with the coefficients given by the selected components 
$\widetilde{C}_{d;jk}$
being zero.
We simply moved some of the terms in
 (\ref{linearConstraintsHomog})
from the right-hand side to the left-hand side. 
\begin{align}
u_{j^{[d]}k^{[d]}}&=
\sum\limits_{j=0}^{D-1}\sum\limits_{k=0}^{n-1} 
\widetilde{C}_{d;jk}u_{jk}
 & d=0\dots \mathrm{rank}(C_{d;jk})-1
\label{constraintsConv}
\end{align}
Then any $u_{jk}$ satisfying the linear constraints (\ref{linearConstraintsHomog})
 can be expressed as a linear combination
of selected components.
Let us denote these components as some general variables
 $V_p$, $p=0\dots N_V-1$
\begin{align}
N_V&=Dn-\mathrm{rank}(C_{d;jk}) \label{Nvvariables} \\
u_{jk}&=\sum\limits_{p=0}^{N_V-1}M_{jk;p} V_p
\label{uExprVp}
\end{align}
Substitute (\ref{uExprVp}) back
to (\ref{EPL}) to obtain an unconstrained generalized eigenproblem (\ref{EPLV})
with respect to
$N_V$ generalized variables $V_p$.
The linear constraints
(\ref{linearConstraintsHomog})
are incorporated into new variables $V_p$,
resulting in a reduction of the total number of variables by the 
 $\mathrm{rank}(C_{d;jk})$.\footnote{
Note that at this stage, some of the vectors that were removed by the constraints can possibly be reintroduced into the
 $V_p$ basis.
They can be directly added as additional column(s) to 
 $M_{jk;p}$.
This changes only the size of the
 $V_p$ basis
to $N_V=Dn-\mathrm{rank}(C_{d;jk})+N_{inj}$.
A better option, however, is to modify $C_{d;jk}$
initially and avoid injecting vectors into the basis.
}
The transformation in 
 (\ref{uExprVp})
is actually a regular
\href{https://en.wikipedia.org/wiki/Gaussian_elimination}{Gaussian elimination},
which is a special form of 
\href{https://en.wikipedia.org/wiki/LU_decomposition}{LU decomposition}.
A simple implementation with row and column pivoting
is used in
\texttt{\seqsplit{com/polytechnik/utils/EliminateLinearConstraints\_HomegrownLUFactorization.java}}.
For a very large number of linear constraints, a transformation such as
\href{https://en.wikipedia.org/wiki/RRQR_factorization}{RRQR factorization}
is probably more numerically stable.
The new eigenproblem involves the matrices
 $\mathcal{S}_{p;p^{\prime}}$
in the numerator and
$Q_{p;p^{\prime}}$ in the denominator.
\begin{subequations}
\label{matrTransf}
\begin{align}
\mathcal{S}_{p;p^{\prime}}&=
\sum\limits_{j,j^{\prime}=0}^{D-1}\sum\limits_{k,k^{\prime}=0}^{n-1}
M_{jk;p}
\mathcal{S}_{jk;j^{\prime}k^{\prime}}
M_{j^{\prime}k^{\prime};p^{\prime}} \label{transfS} \\
Q_{p;p^{\prime}}&=
\sum\limits_{j,j^{\prime}=0}^{D-1}\sum\limits_{k=0}^{n-1}
M_{jk;p}
Q_{jj^{\prime}}
M_{j^{\prime}k;p^{\prime}} \label{transfQ}
\end{align}
\end{subequations}
The $\lambda_{p;p^{\prime}}$ converts in the same way.
We can express eigenproblem
 (\ref{EPL}) in the following form:
\begin{align}
&\frac{
\sum\limits_{p,p^{\prime}=0}^{N_V-1}
V_p
\mathcal{S}_{p;p^{\prime}}
V_{p^{\prime}}
}{
\sum\limits_{p,p^{\prime}=0}^{N_V-1}
V_pQ_{p;p^{\prime}}V_{p^{\prime}}
} 
 \xrightarrow[V]{\quad }\max
 \label{EPLV} \\
 &   \sum\limits_{p^{\prime}=0}^{N_V-1}
S_{p;p^{\prime}}V^{[s]}_{p^{\prime}}
-\sum\limits_{p^{\prime}=0}^{N_V-1} \lambda_{p;p^{\prime}} V^{[s]}_{p^{\prime}}
=\mu^{[s]}\sum\limits_{p^{\prime}=0}^{N_V-1} Q_{p;p^{\prime}} V^{[s]}_{p^{\prime}}
\label{EPLeVV}
\end{align}
The cost of this conversion is that if we originally set
 $Q_{jj^{\prime}}=\delta_{jj^{\prime}}$,
then in the $V_p$ basis, the denominator in (\ref{EPLV})
is no longer a unit matrix.
This is not an issue, as any modern linear algebra package internally converts a generalized eigenproblem to a regular one.
See, for example,
\href{https://www.netlib.org/lapack/lug/node54.html}{DSYGST, DSPGST, DPBSTF},
and similar subroutines.

Let the  original eigenproblem (\ref{EPLeV})
be already solved, and extremal $u^{[s]}_{jk}$ satisfying
the partial quadratic constraint
(\ref{optimmatrixConstraintScalarAppendix}) are obtained.
Consider the Lagrangian variation $\delta \mathcal{L}/\delta u_{sq}$.
If the state $u_{jk}$ is extremal in (\ref{EPL}),
then the variation (\ref{variationU}) is zero
\begin{align}
0&= \sum\limits_{j^{\prime}=0}^{D-1}\sum\limits_{k^{\prime}=0}^{n-1}
S_{sq;j^{\prime}k^{\prime}}u_{j^{\prime}k^{\prime}}
- \sum\limits_{j^{\prime}=0}^{D-1} \breve{\lambda}_{sj^{\prime}} u_{j^{\prime}q}
\label{variationUIngivenState} \\
\breve{\lambda}_{jj^{\prime}}&=\lambda_{jj^{\prime}}+\mu^{[s]} Q_{jj^{\prime}}
\label{lanRenormalized}
\end{align}
It is \textsl{exactly zero} (for all $Dn$ equations)
if $u_{jk}$ is one of the $u^{[s]}_{jk}$
and the Lagrange multipliers are as in (\ref{lanRenormalized}).
However, this $u_{jk}$
may not satisfy the full set of constraints
 (\ref{optimmatrixConstraintAppendixNUDIAG}).
It is adjusted
with (\ref{uAdjEx}) to
satisfy all the constraints, and a new $\lambda_{jj^{\prime}}$
is calculated
to use in the problem (\ref{EPLeV}).
A difficulty we encountered in \cite{malyshkin2019radonnikodym,malyshkin2022machine}
is that this iterative algorithm,
when $\lambda_{jj^{\prime}}$ (\ref{newLambdaSolPartial}) is used,
does not converge to a solution.
In this paper, this difficulty is overcome by solving the eigenproblem
 (\ref{EPLeV})
with additional linear constraints (\ref{linearConstraintsHomog}).
The improved iterative algorithm becomes the following:

\begin{enumerate}
\item
\label{firstStepLambda}
  Take initial $\lambda_{ij}$
  and linear constraints $C_{d;jk}$
to solve the optimization problem (\ref{EPLV})
  with respect to $V_p$.
The solution method involves solving an eigenvalue problem of dimension
$N_V$,
which corresponds to the number of columns in the $M_{jk;p}$ matrix.
A new $u_{jk}$ is obtained from $V_p$ using (\ref{uExprVp}).
The result: $s=0\dots N_V-1$ eigenvalues $\mu^{[s]}$ and corresponding
matrices $u_{jk}^{[s]}$ reconstructed from $V_p^{[s]}$.
The value of $N_V$ is typically $Dn-(D-1)(D+2)/2$.
\item
\label{EVSelectionFromAll}
A heuristic is required
to select the $u_{jk}$  among all $N_V$ eigenstates.
Trying a number of them
and selecting the maximum
(i.e. from all $s=0\dots N_V-1$ select the best state among the top eigenvalues
$\mu^{[s]}$: try all positive and 10 highest negative ones)
providing a large value of the original
functional (\ref{allProjUKxfDIAGGm05})
typically results in only a local maximum.
Numerical experiments show that the index of the eigenstate
is a good invariant:
\textsl{always} selecting the state $V_p^{[s]}$
with the largest $\mu^{[s]}$,
second largest $\mu^{[s]}$,
third largest $\mu^{[s]}$, etc.,
converges to a different solution of the original problem.
Starting with about the fifth largest eigenvalue, convergence may not always be observed.
The global maximum of $\mathcal{F}$ typically corresponds
to selecting the largest $\mu^{[s]}$. A good heuristic
is to run the algorithm $\mathfrak{n}$ times,
always selecting the state of the $\mathfrak{n}$-th largest $\mu^{[s]}$.
Then select the global maximum among these $\mathfrak{n}$ runs;
the remaining $\mathfrak{n}-1$ solutions are also good ---
this way we managed to obtain up to a dozen different solutions.
At worst ---
a cycle without convergence
(usually with a period of $2$ iterations) is observed;
this was an issue in \cite{malyshkin2022machine}.
With the linear constraints technique of Appendix \ref{linConst}
this difficulty is overcome.
\item \label{lagrangeMultipliersStep}
The obtained $u_{jk}$
 is not partially unitary since constraint
 (\ref{optimmatrixConstraintScalarAppendix})
 is a subset of the full constraints (\ref{optimmatrixConstraintAppendixNUDIAG}).
  Apply the adjustment  (\ref{uAdjEx})
  and calculate the $\lambda_{ij}$  (\ref{newLambdaSolPartial})
  in the adjusted state;
  these are the Lagrange multipliers for the next iteration.
\item
For a good convergence, in addition to $\lambda_{ij}$,
we need to select a subspace for the next iteration's variation of
$u_{jk}$.
Using the full size $Dn$ basis
leads to poor convergence \cite{malyshkin2022machine}.
There are two feasible options to improve it:
either use an advanced method for calculating $\lambda_{ij}$,
as detailed in Appendix \ref{LambdaSubspace} below,
or constrain the subspace of $u_{jk}$ variation,
as discussed in Appendix \ref{linConst} below.
The latter technique of additional linear constraints
$C_{d;jk}$ (\ref{linearConstraintsHomog})
provides superior
results.
\item             
    Insert this new $\lambda_{ij}$ into (\ref{lagrangetovariateNUDlenMatrMax})
    and, using the basis $V_p$ obtained (\ref{uExprVp}) from $C_{d;jk}$,
calculate the matrices for the numerator and denominator of the generalized eigenproblem
(\ref{matrTransf})
to be used in the next iteration.
Repeat the iteration process until converging to a maximum (presumably global) of
 $\mathcal{F}$ with $u_{jk}$ satisfying
    the constraints (\ref{optimmatrixConstraintAppendixNUDIAG}).
    If convergence is achieved, the $\lambda_{ij}$ stops changing
    from iteration to iteration, and the $\mu^{[s]}$ of the selected state
    in step \ref{EVSelectionFromAll} above 
    is close to zero.
    On the first iteration, take initial values of Lagrange multipliers $\lambda_{ij}=0$ and have no linear constraints.
\end{enumerate}
If, in the eigenvalue selection step \ref{EVSelectionFromAll}, the state of the maximal eigenvalue is unconditionally selected,
the algorithm becomes linear: it contains no conditional expressions, no selectively executed instructions, no ``branching''.
In this case, it possesses a very simple flat logic that converges to the global maximum of the objective function.
This represents a repeated process of:
    Finding the state corresponding to the maximal eigenvalue of an eigenproblem.
    Adjusting the found state to satisfy the full set of constraints.
    Finding from it the Lagrange multipliers $\lambda_{ij}$ and the constraints $C_{d;jk}$ (a subspace for the next iteration’s variation).
Such a flat structure greatly simplifies the algorithm's complexity analysis,
computer implementation, and offers great potential for parallelization.

Based on a number of numerical experiments, we can conclude that this iterative algorithm almost always converges.
Determining the exact convergence domain is a subject of future research.
A distinctive feature of this algorithm is that instead of the usual iteration internal state in the form of a pair (approximation, Lagrange multipliers) 
$(u_{jk},\lambda_{ij})$,
it uses an iteration internal state in the form of a triple (approximation, Lagrange multipliers, homogeneous linear constraints) 
$(u_{jk},\lambda_{ij},C_{d;jk})$.
Whereas most optimization algorithms use linear system solutions (Newtonian iteration) as a building block, the algorithm in question employs eigenproblem solutions as the building block.
This allows us to develop a much more fine-grained solution selection in step 
 \ref{EVSelectionFromAll} above,
which makes the algorithm less sensitive to degeneracy and more likely to converge to the global maximum.
The drawback of using an eigenproblem solution as a building block is that it is more computationally costly than a linear system solution or first-order gradient methods.
However, the main goal of the paper is to present a working proof-of-concept algorithm capable of solving a new algebraic
 ``eigenoperator'' problem (\ref{eigenvaluesLikeProblem});
the computational complexity of optimization is a separate concern.
An apparent optimization would be to replace a general-purpose eigenproblem solver with one that finds only the largest eigenvalue in step \ref{EVSelectionFromAll}, which is typically adequate and expected to significantly enhance algorithm performance,
see Appendix \ref{compComplexityAnalysis} for a preliminary analysis.

\begin{table*}
\caption{\label{comparisonConvergence}
Comparison of convergence between the algorithm presented in this paper and the one from our previous work \cite{malyshkin2022machine},
conducted on a data sample 
with $D=4$, $n=19$, $M=13540$
for the first $i=0\dots 17$ iterations.
We present: $\mu$ -- the eigenvalue of the
\hyperref[EVSelectionFromAll]{selected} state,
$\mathcal{F}$ -- the value of the  functional (\ref{allProjUKxfDIAGGm05})
\hyperref[normalizingToNumberObservations]{normalized} to the number of observations,
and an
\hyperref[penaltyFun]{indicator of unitarity} $\sum 1/\lambda_G$.
}
\begin{ruledtabular}
\npdecimalsign{.}
\nprounddigits{2}
\npfourdigitnosep
\begin{tabular}{r|n{5}{2}n{5}{2}n{5}{2}|n{5}{2}n{5}{2}n{5}{2}}
 &\multicolumn{3}{c|}{KGOIterationalSubspaceLinearConstraints}&\multicolumn{3}{c}{KGOIterationalSimpleOptimizationU}\\
i & \multicolumn{1}{c}{$\mu$} & \multicolumn{1}{c}{$\mathcal{F}$} &  \multicolumn{1}{c|}{$\sum 1/\lambda_G$}
   & \multicolumn{1}{c}{$\mu$} & \multicolumn{1}{c}{$\mathcal{F}$} &  \multicolumn{1}{c}{$\sum 1/\lambda_G$} \\
\hline
0 & 2296.9664381293956   & 7864.08208568746  & 38.85764636202683
 & 2296.9664381293946   & 7864.082085687466  & 38.857646362026486 \\
1 & 201.08292423140773    & 7936.317627722794  & 20.017895600080813 
 & 239.83937285483216    & 8020.582098643324  & 34.939165999833364 \\
2 & 113.9767306161277   & 8010.69429163704  & 27.038276189371903 
 & 110.62455788990813   & 8032.885989077905  & 17.69637116047035 \\
3 & 100.82020149761865    & 8178.92235535865  & 7.073880638170096 
 & 53.941772755344715   & 8014.302236229355  & 14.310608565633178 \\
4 & 64.90956164872827    & 7890.372439650525  & 26.06242106324566 
 & 150.5958980440506    & 8138.494564158976  & 15.822630864304209 \\
5 & 204.70518316408067   & 8078.925356460416  & 32.47103039281571 
 & 57.32929410041896   & 8113.737629073397  & 11.394380490141037 \\
6 & 134.2406441579564   & 7873.865422253332  & 34.30027863305253 
 & 109.04733186020607   & 8071.2425713607145  & 17.89561571030589 \\
7 & 201.3851590207987   & 8112.941414698682  & 24.75629431991697 
 & 38.23758734479807    & 8005.977039865902  & 11.930285516225728 \\
8 & 90.7813367762079   & 7869.540956852294  & 34.37528792609749 
 & 92.90960438819266   & 8059.795821350578  & 18.311503292481955 \\
9 & 159.770324242123   & 8057.352366071262  & 12.583322607042625 
 & 49.791918171772494   & 8101.236318979962  & 6.5110920167446675 \\
10 & 83.14537969420344   & 8212.672220428809  & 8.062257216355384 
 & 148.46434284267005   & 8094.143293812645  & 29.023280432811003 \\
11 & 28.154310093878394   & 8194.824362989539  & 6.2200697896128165 
 & 86.10883419509504    & 8035.4113798035305  & 16.244174235871377 \\
12 & 25.062479073281953  & 8292.054504699776  & 4.03408204351225 
 & 156.60097602668628  & 8060.7740424468875  & 26.22032155311293 \\
13 & 3.0765070718083045    & 8302.515628058409  & 4.021321809862028 
 & 95.36774881700336   & 8112.066528030416  & 17.71736839959328 \\
14 & 0.2322107927923612    & 8303.455070601445  & 4.000014107729678 
 & 73.72947951341787   & 8016.439620023038  & 23.42292376673859 \\
15 & 5.627615039432662E-4   & 8303.457323791197  & 4.00000000035633 
 & 202.11481988567772   & 8040.93154248803  & 32.04671507208003 \\
16 & 1.0538950549524283E-8    & 8303.457323833356  & 3.999999999999999 
 & 93.91219490729446    & 7963.836816487771  & 21.198581432963753 \\
17 & 2.392928545221896E-13   & 8303.457323833352  & 4.0 
 & 192.01644524589113   & 8030.496455853109  & 24.14459878733836 \\
\end{tabular}
\end{ruledtabular}
\end{table*}

A reference implementation of this algorithm is available
at \texttt{\seqsplit{com/polytechnik/kgo/KGOIterationalSubspaceLinearConstraints.java}}.
There are dozens of other algorithms implemented, but only this one provides iterative convergence.\footnote{
An equivalent implementation with additional features\cite{belov2024quantum},
\texttt{\seqsplit{com/polytechnik/kgo/KGOIterationalSubspaceLinearConstraintsB.java}},
follows the same logic, but all calculations are performed in the original basis without using $v^{[s]}_{jk}$. 
The result is identical to \texttt{\seqsplit{KGOIterationalSubspaceLinearConstraints.java}},
which once again demonstrates the basis-invariance of our theory.
}
There is a driver in the \texttt{\seqsplit{com/polytechnik/kgo/KGOSolutionVectorXVectorF.java:main}} class
for a quick test of the algorithm.
The driver generates a deterministic random sample
from which the tensor $S_{jk;j^{\prime}k^{\prime}}$ is calculated,
and then the specified algorithm (taken from the command argument) is applied to find the optimal partially unitary operator.
Let us compare the convergence of this paper's iterative algorithm with the results from \cite{malyshkin2022machine}
(that uses vanilla Lagrange multipliers) on the driver's deterministically randomly generated data with default settings: $D=4$ and $n=19$ of $M=13540$ observations.\\
\texttt{\seqsplit{java\ com/polytechnik/kgo/KGOSolutionVectorXVectorF\ ITERATIONS\_KGOIterationalSubspaceLinearConstraints\ 2>\&1\ |\ grep\ Selected}}\\
and \\
\texttt{\seqsplit{java\ com/polytechnik/kgo/KGOSolutionVectorXVectorF\ ITERATIONS\_KGOIterationalSimpleOptimizationU\ 2>\&1\ |\ grep\ Selected}}\\
The output is \texttt{grep}-filtered to show only the iteration status.
The results are presented in Table \ref{comparisonConvergence}.
From the table, we see that the algorithm cycles through a number of initial iterations without convergence.
However, once it enters a
\href{https://en.wikipedia.org/wiki/Contraction_mapping}{contraction mapping}
area, convergence becomes very fast (faster than any geometric progression).
The plain vanilla Lagrange multipliers method does not converge at all:
optimizing (\ref{EPL}) over the full $Dn$ space
severely violates constraints  (\ref{optimmatrixConstraintAppendixNUDIAG}),
and problem degeneracy impedes convergence.
This can be demonstrated by running
\texttt{\seqsplit{KGOIterationalSubspaceLinearConstraints}}
in the standard manner for $20$ iterations to obtain a perfect solution. 
Then, starting from iteration 21, turn off the linear constraints
(\ref{constraintsCcoefs}), effectively switching to the vanilla Lagrange multipliers method internally.
The result is that, for the next 4 to 7 iterations, the solution remains almost the same. However, after 7-10 iterations, it starts to diverge due to accumulated floating-point errors, leading to irregular cycling without convergence.
This demonstrates the critical importance of linear constraints
(\ref{constraintsCcoefs})
for the convergence of iterative algorithms and the intrinsic instability of vanilla methods.

There are several QCQP software packages available, both commercial and open-source, such as 
\cite{frison2022introducing},
\href{https://github.com/cvxgrp/qcqp}{python qcqp},
\href{https://nag.com/solving-quadratically-constrained-quadratic-programming-qcqp-problems/}{NAG QCQP}
among many
\href{https://en.wikipedia.org/wiki/Quadratically_constrained_quadratic_program#Solvers_and_scripting_(programming)_languages}{others}.
A question arises: How well can existing software handle a QCQP optimization problem like that in this paper?
We did not extensively test third-party software (although we plan to do so in the future), but generally, the results were unsatisfactory, especially in terms of finding the global maximum.
The reason is that existing QCQP software packages were designed as general-purpose solvers intended to solve any QCQP problem. Many of them are heuristic Newton-style solvers with Lagrange multipliers often combined with some form of linear programming and convex optimization.
The optimization problem considered in this paper is special: it has only quadratic equality constraints, is non-convex, and exhibits multiple solutions, local maxima, and multiple saddle points.
For such a problem, solvers based on single solution iterations (e.g., Newtonian type (second order), gradient type (first order), etc.) are unlikely to identify the global maximum.
One may try
\href{https://en.wikipedia.org/wiki/Genetic_programming}{Genetic programming}
solvers, but they lack knowledge about the underlying algebraic structure of the problem.
The problem of finding an operator that optimally converts an operator from one Hilbert space to another requires a specialized solver.
Our use of the generalized eigenvalue problem as the algorithm's building block has the advantage of obtaining many solution candidates (eigenvectors) at once. With proper
\hyperref[EVSelectionFromAll]{selection},
we can greatly increase the chances of finding the global maximum.
Compared to \cite{malyshkin2022machine}, the quality of our optimization algorithm has been greatly improved, and we intend to use it as a ``black box'' for applying it to several problems for demonstration purposes.
Let us demonstrate the value of finding a partially unitary operator that optimally transforms a vector
from Hilbert space
\emph{IN} to Hilbert space \emph{OUT}.

\section{\label{recoverUnitaryMatrix}A Demonstration Of Recovering Unitary Dynamics From Phase-Stripped Data}

Consider an inverse problem.
Let a dynamic system evolve with the equation 
$\mathbf{X}^{(l+1)}=\mathcal{U}\mathbf{X}^{(l)}$
(\ref{unitaryDynamcs}).
The problem is to
recover the orthogonal operator $\mathcal{U}$ from an observed sample
$\mathbf{X}^{(l)}$.
The problem would be trivial
if $\mathbf{X}^{(l)}$
were directly observed --- a regression analysis could reveal $\mathcal{U}$,
see e.g. the section
``An application of LRR representation solution to dynamic system identification
problem'' of \cite{malyshkin2019radonnikodym}.
What greatly complicates matters is that we consider observations of a quantum channel type; 
hence,  $\mathbf{X}^{(l)}$ is observable only up to a phase.
This means that any wavefunction mapping
(\ref{mlproblemVector})
can only be observed with an
unknown phase.
This can be modeled by multiplying the actually measured classical system values by
\hyperref[fnRandomPhase]{random phase factors}.
The problem is to determine the operator $\mathcal{U}$
from the sample given in (\ref{timeEvolutionSample}).\footnote{
 The operator itself
is defined within a phase.
Further degeneracy may arise from the data; for instance, if some components 
$X_j$ and $X_{j+1}$
consistently remain the same in the sample, then we cannot distinguish between a unit matrix and a permutation matrix.
The input data sample should be information-complete (IC) \cite{torlai2023quantum}.
}

In this section a few simple examples are presented. All calculations are
performed in real space, using a $\pm 1$ factor instead
of a complex phase factor.
A timeserie is initially generated with equation (\ref{unitaryDynamcs}).
The obtained vectors $\mathbf{X}^{(l)}$ are then multiplied by random $\pm 1$  factors.
These new vectors (\ref{timeEvolutionSample}) are now considered as observables.
The problem is to recover $\mathcal{U}$ from observable data.
We present a recovery with two quantum channels: transforming the Gram matrix of
Section \ref{UnitaryLearning}
and transforming the unit matrix of
Section \ref{traditionalUL}.
Since the unitarity of the test data is exact,
both methods recover the underlying operator $\mathcal{U}$ exactly.
Therefore, only the Gram matrix quantum channel is presented below.

Let us start with a simple
\href{https://en.wikipedia.org/wiki/3D_rotation_group}{SO(3)}
rotation group. Rotation  matrices can be represented
using
\href{https://en.wikipedia.org/wiki/Euler_angles}{Euler angles}
$(\varphi,\theta,\psi)$ (in a space of dimension $d$,
there are a total of $d(d-1)/2$ angles).
\begin{align}
g(\varphi,\theta,\psi)&=
\begin{pmatrix}
\cos \varphi & -\sin \varphi & 0 \\
\sin \varphi & \cos \varphi & 0 \\
0 & 0 & 1 
\end{pmatrix}
\begin{pmatrix}
1 & 0 & 0  \\
0 & \cos \theta & -\sin \theta  \\
0 & \sin \theta & \cos \theta 
\end{pmatrix} \nonumber \\
&\times
\begin{pmatrix}
\cos \psi & -\sin \psi & 0 \\
\sin \psi & \cos \psi & 0 \\
0 & 0 & 1 
\end{pmatrix}
\label{rotMatrix}
\end{align}
Take $\mathcal{U}=g(\varphi=0.1,\theta=0.4,\psi=0.7)$
\begin{align}
\npdecimalsign{.}
\nprounddigits{4}
\npfourdigitnosep
\mathcal{U}=
\begin{pmatrix}
\numprint{0.7017836283209663} & \numprint{-0.7113285603155088} & \numprint{0.03887696361761665} \\
\numprint{0.6667562447219523} & \numprint{0.6366324552659786} & \numprint{-0.38747287263277136}\\
\numprint{0.2508701838500143} & \numprint{0.2978435767000479} & \numprint{0.9210609940028851}
\end{pmatrix}
\end{align}
and apply it to transform the initial vector 
$\npdecimalsign{.}
\nprounddigits{4}
\npfourdigitnosep
(\numprint{0.09205746178983236}, \numprint{0.5523447707389941}, \numprint{0.8285171561084912})$
through 1000 transformations.
The datafile \texttt{SO3.csv}
is generated using \\
\texttt{\seqsplit{java\ com/polytechnik/algorithms/PrintOrthogonalSeq}} \\
\texttt{
Writing 1000 points to /tmp/SO3.csv
}. \\
There is a simple demo program that recovers operator $\mathcal{U}$
from sampled data (\ref{timeEvolutionSample})
with random phases $\pm 1$ possibly introduced into observations.
This program maximizes (\ref{classicUnitaryDynamics})
with a crude regularization (\ref{nBasis})
(numerical stability depends on regularization,
but the optimization result does not \cite{malyshkin2019radonnikodym})
and does not address possible data degeneracy.
However, its simplicity makes its internals clear.
\\
\texttt{\seqsplit{java\ com/polytechnik/algorithms/DemoRecoverUnitaryMatrixFromSeq\ --data\_file\_to\_build\_unitarymodel\_from=/tmp/SO3.csv\ --data\_cols=6:1,3:0}}\\
The program recovers the $\mathcal{U}$ identically
\begin{align}
\npdecimalsign{.}
\nprounddigits{4}
\npfourdigitnosep
\mathcal{U}=
\begin{pmatrix}
\numprint{-0.7017836283209665} & \numprint{0.7113285603155112} & \numprint{-0.03887696361761664} \\
\numprint{-0.6667562447219517} & \numprint{-0.63663245526598} & \numprint{0.38747287263277136} \\
\numprint{-0.25087018385001414} & \numprint{-0.29784357670004835} & \numprint{-0.9210609940028854} \\
\end{pmatrix}
\label{SO3recovered}
\end{align}
As discussed earlier, the operator $\mathcal{U}$ is defined within a
$\pm 1$ factor in real space.
The same test was run on randomly generated orthogonal
matrices of dimensions $3,5,7,17,40$. All tests can be run automatically using \\
\texttt{\seqsplit{java\ com/polytechnik/algorithms/PrintOrthogonalSeq\textbackslash \$TestAuto}}\\
The maximal difference in matrix elements
(for all dimensions tried)
is less that $10^{-13}$,
which is about floating point errors.
This leads us to conclude that the developed numerical algorithm in
Section \ref{numericalSolutionSect}
can recover system dynamics from wavefunction measurements
(without a phase) for dynamic systems of high dimension.
This algorithm is a powerful method for solving the inverse problem in quantum mechanics.

\section{\label{polynomialMapping}A Demonstration Of Polynomial Mapping Recovery}
Consider polynomials in the $[-1:1]$ interval.
Let there be $l=1\dots M$ points $y^{(l)}$ split equidistantly in the interval.
Define the data
\begin{subequations}
\label{polynXF}
\begin{align}
x_k^{(l)}&=\xi^{(l)}T_k(y^{(l)}) & \text{$k=0\dots n-1$}  \label{Xpol} \\
f_j^{(l)}&=\zeta^{(l)}P_j(y^{(l)}) & \text{$j=0\dots D-1$} \label{Fpol}
\end{align}
\end{subequations}
Here, $T_k$ is
a \href{https://en.wikipedia.org/wiki/Chebyshev_polynomials}{Chebyshev polynomial}
and $P_j$ is 
a \href{https://en.wikipedia.org/wiki/Legendre_polynomials}{Legendre polynomial}.
The $\xi^{(l)}$ and $\zeta^{(l)}$ are deterministic random functions
of $l$ that take the value $\pm 1$.
The problem is to build the $u_{jk}$ matrix that maximizes (\ref{allProjUKxfAppendix})
subject to the constraints in 
(\ref{optimmatrixConstraintAppendixNUAS}).
The mapping (\ref{fProjXDifferently}) for the (\ref{polynXF})
data has the meaning of finding the coefficients that expand $D$ Legendre polynomials in $n$ Chebyshev polynomials.
The problem would be trivial
(reduced to solving a linear system)
were it not for the random
$\pm 1$ factors $\xi^{(l)}$ and $\zeta^{(l)}$ in (\ref{polynXF}).
With the presence of random phases,
any direct projection of one basis onto another becomes unavailable.
The only feasible way is to consider the quantum channel mapping
(\ref{fProjXDifferently}).
The solution is similar to that of the previous section.
Calculate the Gram matrices 
 (\ref{GramX}), (\ref{GramF})
and the $S_{jk;j^{\prime}k^{\prime}}$ tensor (\ref{SfunctionalClassic}).
After any regularization of the input data, for example (\ref{nBasis}),
apply the algorithm from Section \ref{numericalSolutionSect}.
Specifically, generate
a data file named
 \texttt{ChebyshevLegendre.csv}
by running \\
\texttt{\seqsplit{java\ com/polytechnik/algorithms/PrintChebyshevToLegendreMapping}}\\
\texttt{
Writing 500 points to \texttt{\seqsplit{/tmp/ChebyshevLegendre.csv}}
} \\
This file contains $500$ points in the interval 
 $[-1:1]$ of (\ref{polynXF}) data,
with $T_0\dots T_{10}$ and $P_0\dots P_5$ randomly multiplied by $\pm 1$ factors.
Consider the simple task of converting $T_0\dots T_4$ to $P_0\dots P_4$.
Use a simple demonstration program
that recovers the mapping (\ref{fProjXDifferently})
from sampled data (\ref{polynXF}), with random phases $\pm 1$
possibly introduced into observations (the approach is phase-agnostic).
This program maximizes $\mathcal{F}$, and its
simplicity makes its internals clear. \\
\texttt{\seqsplit{java\ com/polytechnik/algorithms/DemoRecoverMapping\ --data\_file\_to\_build\_model\_from=/tmp/ChebyshevLegendre.csv\ --data\_cols=21:2,6:14,18:-1:0}}\\
The program recovers the mapping identically (values below $10^{-15}$ are replaced by zero).
\begin{align}
&\mathcal{U}= \label{PTmapping} \\
&\npdecimalsign{.}
\nprounddigits{1}
\npfourdigitnosep
\npthousandthpartsep{}
\begin{pmatrix}
\numprint{-1.0} & 0 & 0 & 0 & 0 \\
0 & \numprint{-1.0000000000000007} & 0 & 0 & 0 \\
\nprounddigits{2}\numprint{-0.24999999999999944} & 0 & \nprounddigits{2}\numprint{-0.7499999999999999} & 0 & 0 \\
0 & \nprounddigits{3}\numprint{-0.37500000000000056} & 0 & \nprounddigits{3}\numprint{-0.625} & 0 \\
\nprounddigits{6}\numprint{-0.1406250000000003} & 0 & \nprounddigits{4}\numprint{-0.3125000000000007} & 0 & \nprounddigits{6}\numprint{-0.5468750000000009}
\end{pmatrix}
\nonumber
\end{align}
Exact values can be obtained using a polynomial library from \cite{2015arXiv151005510G},
which is included with the
\href{http://www.ioffe.ru/LNEPS/malyshkin/code_polynomials_quadratures.zip}{software for this paper}\cite{polynomialcode}.
Run in jshell \texttt{new Legendre().convertBasisToPBASIS(5,new Chebyshev())}
to obtain $P_j$ over $T_k$ expansion
\begin{align}
\begin{pmatrix}
 1.0 & \\
 0 & 1.0  \\
 0.25 & 0 & 0.75 \\
 0 & 0.375 & 0 & 0.625 \\
 0.140625 & 0 & 0.3125 & 0 & 0.546875
 \end{pmatrix}
 \label{polExpansionOrig}
\end{align}
i.e. $P_3=0.375 T_1 +0.625 T_3$. This matches (\ref{PTmapping}) within a $\pm$ sign.
In this  $n=D$ case we have a perfect recovery.
Now consider the same problem of polynomial mapping
for $D<n$, let us run it with $D=4$ and $n=5$.\\
\texttt{\seqsplit{java\ com/polytechnik/algorithms/DemoRecoverMapping\ --data\_file\_to\_build\_model\_from=/tmp/ChebyshevLegendre.csv\ --data\_cols=21:2,6:14,17:-1:0}}\\
The obtained mapping
\begin{align}
&\mathcal{U}= \label{umappingNotTracePreserving}\\
&\npdecimalsign{.}
\nprounddigits{5}
\npfourdigitnosep
\npthousandthpartsep{}
\begin{pmatrix}
0 & \numprint{-1.2943396399454876} & 0 & \numprint{0.5602251187496122} & 0 \\
\numprint{-0.6394960219645237} & 0 & \numprint{-0.44947631527213006} & 0 & \numprint{0.07596007449430658}\\
0 & \numprint{-0.6767757244915493} & 0 & \numprint{-0.6927388497867077} & 0 \\
\numprint{-0.14966949387605982} & 0 & \numprint{-0.6208267307592447} & 0 & \numprint{-0.5023291751872818}\\
 \end{pmatrix}
 \nonumber
\end{align}
is not a subset of the $D=5$, $n=5$ mapping (\ref{PTmapping}),
a subset can be obtained by running with $D=4$, $n=4$.
We previously discussed \cite{malyshkin2022machine}
the difficulties of the $D<n$ case.
A mapping between two Hilbert spaces of different dimensions
sometimes leads to unusual behavior, as the mapping is no longer trace-preserving.
The value of $\mathcal{F}$ increases
when going from $D=4$, $n=4$ to $D=4$, $n=5$ but the mapping is no longer a subset.
Consider a simple numerical experiment demonstrating the behavior of trace-decreasing maps.

\begin{figure}
\includegraphics[width=0.95\columnwidth]{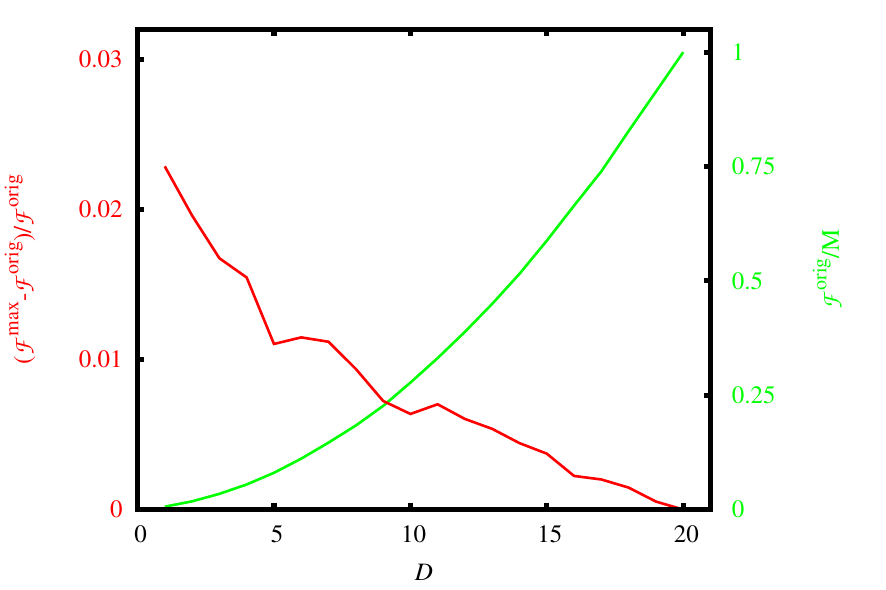}
\caption{\label{FIgFidelityMax}
For $D\le n=20$,
a sample
$\psi_l\to\phi_l$ 
is constructed with a known $u_{jk}$,
the ``original'' one.
The result is compared with the fidelity-maximizing solution
of optimization problem (\ref{allProjUKxfAppendix}).
For $D<n$ (partial unitarity),
the fidelity of the optimization problem solution
is always greater than the original fidelity.
For $D=n$ (unitarity),
the original and optimization problem solutions match exactly.
A deviation of $\mathcal{F}^{orig}/M$ from $1$
 indicates that the transform does not preserve trace
---
 it is a
\hyperref[operatorTransform]{trace preserving map}
only for $D=n$.
}
\end{figure}
\section{\label{funDemoNcNx}A Demonstration of Partially Unitary Behavior $D<n$}
Consider a numerical experiment with a partially unitary mapping where
 $D<n$.
For demonstration we select a unitary matrix
with matrix elements $U_{jk}$ $j,k=0\dots n-1$ and generate a sample
of random $\psi_l$
(each of dimension $n$) with
$M=1000$ observations.
Then, for a given $D$,
we select the first $D$ rows of $U_{jk}$
and use them as the partially unitary operator
$u_{jk}$, $j=0\dots D-1$, $k=0\dots n-1$.
Applying it to every $\psi_l$, a corresponding $\phi_l$
(of dimension $D$) is created,
and the mapping (\ref{mlproblemVector}) is obtained for all sample elements.
Then $S_{jk;j^{\prime}k^{\prime}}$ is created from the sample
and the numerical algorithm of Section \ref{numericalSolutionSect}
is applied. For $D=n$,
this is essentially the same problem considered
in Section \ref{recoverUnitaryMatrix} above.
Now we run the problem for all $D=1\dots n$.
For every $D$, we calculate the original $u_{jk}$ fidelity
and the fidelity of the obtained $u_{jk}$
 as the solution to the optimization problem,
run \texttt{\seqsplit{java\ com/polytechnik/algorithms/DemoDMUnitaryMappingTest\ 2>\&1\ |\ grep\ Map:}}.
In the $D=n$ case both fidelities are identical as well as the operators $u_{jk}$.
For $D<n$, however, the fidelity $\mathcal{F}^{\max}$
calculated with the fidelity-maximizing operator
is greater than the fidelity $\mathcal{F}^{orig}$ of the original
operator used to construct the sample (\ref{mlproblemVector}).
In Fig. \ref{FIgFidelityMax} we present the result for $n=20$ ---
the original operator fidelity (normalized to sample size $M=1000$) and
its relative difference
$(\mathcal{F}^{\max}-\mathcal{F}^{orig})/\mathcal{F}^{orig}$.
We determined that for $D<n$,
$\mathcal{F}^{\max}$ is always greater than $\mathcal{F}^{orig}$.
The behavior is determined by the form of fidelity.
For  fidelity definition (\ref{allProjUKxfAppendix}),
the optimization problem solution in the case of partial unitarity
$D<n$
produces greater fidelity than the fidelity
of the mapping used to construct the sample.
Consider the case where $D=1$.
From the constraints (\ref{partialUnitarityOrtBasis}) it follows
that $\sum_{k=0}^{n-1} u_{0k}^2=1$
and the wavefunction transform is $f_0=\sum_{k=0}^{n-1} u_{0k} x_k$.
The vector $\mathbf{f}$ (now a vector of a single element)
has a unit norm only if $x_k=u_{0k}$ (for $D=1$, this is actually a projection).
For $D<n$, the quantum channel (\ref{operatorTransform})
is a
\href{https://en.wikipedia.org/wiki/Quantum_channel}{trace decreasing map},
and the algorithm finds a mapping with higher fidelity than the original
mapping used to construct the quantum channel in the first place.

\section{\label{funDemo}A Demonstration Of Function Interpolation}
In the two previous sections,
we considered examples of constructing
$u_{jk}$
when the data of $\mathbf{x}\to\mathbf{f}$ form
has vectors $\mathbf{x}$ and $\mathbf{f}$
already belonging to corresponding Hilbert spaces.
For classical measurements, this is often not the case;
a transformation $\mathbf{x}\to\psi$, $\mathbf{f}\to\phi$
is required to build Hilbert space states $\psi$ and $\phi$
from the original observations $\mathbf{x}$ and $\mathbf{f}$.
The most straightforward method to construct such states
is to consider states localized at given $\mathbf{x}$ (or $\mathbf{f}$)
as wavefunctions (\ref{psiYlocalized}) and use them for the mapping
(\ref{mlproblemVectorCladssicProblemPsi}).
As considered in section \ref{classicMapping} above,
localized states $\psi_{\mathbf{y}}(\mathbf{x})$
use the same Gram matrix (\ref{GramX}) for the scalar product,
which is obtained from $l=1\dots M$ sampled data.
Effectively this
is sample average (\ref{gfaverageDef})
with weight multiplied by a
localized at $\mathbf{y}$
non-negative
function $\psi^2_{\mathbf{y}}(\mathbf{x})$ (it is
normalized as $1=\Braket{\psi^2_{\mathbf{y}}(\mathbf{x})}$).
It provides a Radon-Nikodym approximation of some characteristic $g$
\begin{align}
g_{RN}(\mathbf{y})&= \Braket{g\psi^2_{\mathbf{y}}}
\label{approxRN}
\end{align}
The approximated value here is a superposition of the observed
$g$
with the positive density $\psi^2_{\mathbf{y}}(\mathbf{x})$.
The familiar least squares interpolation (\ref{fxapproxLS}),
which expands the value rather than representing a probability,
has a similar form.
\begin{align}
g_{LS}(\mathbf{y})&= \psi_{\mathbf{y}}(\mathbf{y})\Braket{g\psi_{\mathbf{y}}}
\label{approxLS}
\end{align}
Here, the average is taken with the density $\psi_{\mathbf{y}}(\mathbf{x})$ ---
it is not always positive as $\psi^2_{\mathbf{y}}(\mathbf{x})$
in the Radon-Nikodym (\ref{approxRN}) expression;
the $\psi_{\mathbf{y}}(\mathbf{y})$ is a normalizing factor of
$1/\sqrt{K(\mathbf{y})}$.
The expressions \ref{approxRN}) and (\ref{approxLS})
differ in how they represent the
\href{https://en.wikipedia.org/wiki/Dirac_delta_function}{delta function $\delta(\mathbf{x}-\mathbf{y})$}:
as $\psi^2_{\mathbf{y}}(\mathbf{x})$ or as
 $\psi_{\mathbf{y}}(\mathbf{y})\psi_{\mathbf{y}}(\mathbf{x})$.

Consider the scalar function interpolation problem
in the form of mapping between two Hilbert spaces with $u_{jk}$.
Let there be a scalar $f(x)$ and  $M$ observation points
$f^{(l)}=f(x^{(l)})$, $l=1\dots M$.
Convert this scalar mapping to a vector one
(the $\xi^{(l)}$ and $\zeta^{(l)}$ are deterministic random functions
on $l$ that take the value $\pm 1$)
\begin{subequations}
\label{xfScalarVector}
\begin{align}
x^{(l)}_k&= \xi^{(l)} \left(x^{(l)}\right)^k & k=0\dots n-1 \label{xScalarVector} \\
f^{(l)}_j&= \zeta^{(l)} \left(f^{(l)}\right)^j & j=0\dots D-1 \label{fScalarVector}
\end{align}
\end{subequations}
For numerical stability it is better,
instead of monomials (powers of the argument),
to use Chebyshev or Legendre polynomials as
$x^{(l)}_k=\xi^{(l)} T_k(ax^{(l)}+b)$ and $f^{(l)}_j=\zeta^{(l)}T_j(cf^{(l)}+d)$ with $a,b,c,d$ chosen
to bring the argument into $[-1:1]$ interval.
This greatly increases the numerical stability of calculations.
However, the result itself is invariant with respect
to the polynomial basis choice ---
the result will be the same with any polynomial basis.
From the obtained vector to vector mapping construct
$\mathbf{x}^{(l)}$ and $\mathbf{f}^{(l)}$
localized states
$\psi_{\mathbf{x}^{(l)}}(\mathbf{x}) \to \psi_{\mathbf{f}^{(l)}}(\mathbf{f})$
mapping (\ref{mlproblemVector}) with the former one defined
in (\ref{psiYlocalized})
and the latter one obtained from it with 
argument and index replacement (\ref{mlproblemVectorCladssicProblemPsi}).
Put them into (\ref{probgUypkAppAS}) and obtain (\ref{Sclassic}).
After solving the problem for $u_{jk}$ an evaluation of interpolated $f(x)$
can be performed as follows: From a given $x$ construct
the vector $\mathbf{x}$ (\ref{xScalarVector}).
Substitute it to (\ref{probFXUpExpanded}) to obtain the probability
of a given vector of outcome $\mathbf{f}$. In the scalar case
the outcome value can be evaluated, for example,
by constructing (\ref{fScalarVector}) vector $\mathbf{f}$ from $f$
and then finding the value of $f$ that provides the maximal
value of probability (\ref{probFXUpExpanded}). For scalar $f$
this problem can be reduced to finding the roots of a single variable polynomial.

\begin{figure}
\includegraphics[width=0.95\columnwidth]{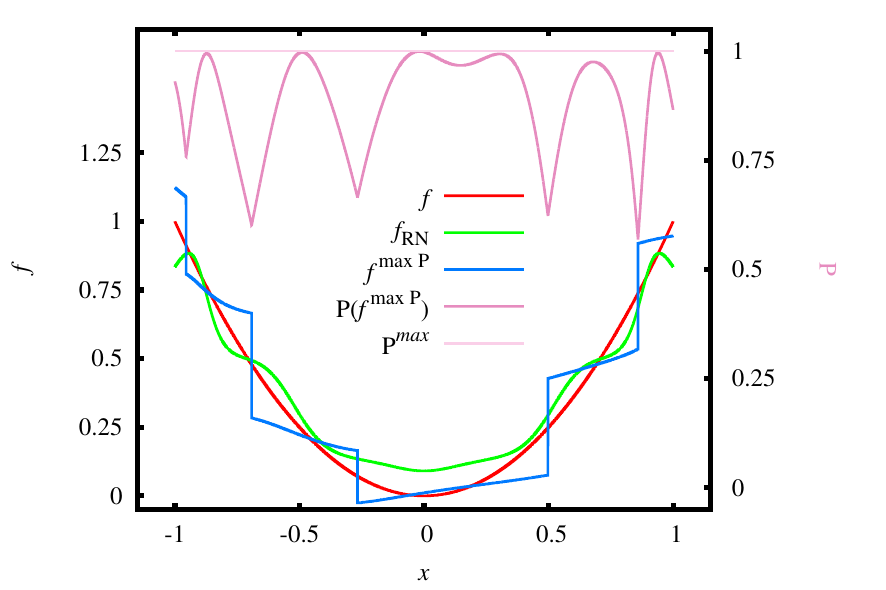}
\caption{\label{ScalarFunInterpolation}Scalar function $f=x^2$ (red)
interpolation with:
green: Radon-Nikodym (\ref{approxRN}),
blue: $f$ corresponding to the maximal  probability (\ref{probFXUpExpanded}),
the dependence has discontinuities;  some numerical instability
presents even in this $D=n=6$ case.
Pink: probabilities corresponding to (\ref{probFXUpExpanded})
and (\ref{f_Pmax}).
}
\end{figure}

This approach creates a number of issues.
An attempt to fit a scalar function $f(x)$
to Hilbert spaces mapping using moments like
$\Braket{f^j x^k |K(\mathbf{x}) K(\mathbf{f})| f^{j^{\prime}} x^{k^{\prime}}}$
(\ref{ChristoffelfunctionsProductMoments})
create difficulties both in computations and in mapping back
from Hilbert space to function value.
As a simple demonstration,
let $f(x)=x^p$ with $p=2$.
This requires calculating  $x$-moments of maximal degree
$2(n-1)+2p(D-1)$,
which for $n=D=6$ gives a maximal polynomial degree of $30$,
creating numerical instability difficulties.
There is an additional
problem of converting back from the Hilbert space to
the value of $f$ by checking all extrema of (\ref{probFXUpExpanded})
to find the  $f$ that provides the maximal probability is a ``switching''
function that creates a non-continuous solution when polynomial roots change.
An example is presented in Fig. \ref{ScalarFunInterpolation}.
For $f=x^2$ (red line),
we calculated the Radon-Nikodym approximation (\ref{approxRN})
(least squares is identical to the original  $f$  since $x^2$ is in the basis,
thus it is not presented in the plot), as well as $f$
corresponding to the maximum probability (blue line)
and its corresponding probability (pink).
In the plot, one can see starting numerical instability (asymmetry in the plot)
and discontinuity in $f$ corresponding to the change of the selected root.
As expected for the $x$ where $P(f^{\max P}(x))=1$ the value of
$f^{\max P}$ is equal to the exact $f(x)$.

Alternatively\cite{malyshkin2022machine},
we can consider (\ref{probFXUpExpanded})
without the requirement (\ref{fScalarVector})
that all components of $\mathbf{f}$ are obtained from a single scalar $f$.
\begin{subequations}
\begin{align}
f^{\max P}_j&=
\sum\limits_{j^{\prime}=0}^{D-1} G^{\mathbf{f}}_{jj^{\prime}}a_{j^{\prime}}
\label{f_Pmax} \\
P(\mathbf{f}^{\max P})\Big|_{\mathbf{x}}&=\sum\limits_{j,j^{\prime}=0}^{D-1}
a_{j}G^{\mathbf{f}}_{jj^{\prime}}a_{j^{\prime}}
\label{PmaxValue}
\end{align}
\end{subequations}
The maximal value of the probability (\ref{probFXUpExpanded})
at (\ref{PmaxValue})
is an important characteristic of the obtained solution.
If the problem dimensions are balanced -- the
value is equal to $1$ for all $x$ (pink line $P^{max}$
in Fig. \ref{ScalarFunInterpolation}),
it can be lower in the case where $D<n$, which
requires separate consideration.
In \cite{malyshkin2022machine} we evaluated $f(x)$
from the first two elements of the vector $\mathbf{f}$:
for a polynomial basis  $Q_j(f)$
(in  (\ref{fScalarVector}) $Q_j(f)=f^j$)
we can obtain the value of $f$ from the ratio
 $Q_0=f^{\max P}_0$ and $Q_1(f)=f^{\max P}_1$.
Back in \cite{malyshkin2022machine},
we believed that observed singularities
were caused by a poor solution to the optimization problem.
While the algorithm developed in this paper provides
a very good solution to the optimization problem
of Hilbert spaces mapping,
the method of obtaining the value of the scalar
$f$
from the ratio of the first two basis functions does not work well.

Currently we do not have a good solution to the problem of converting
two Hilbert spaces mapping $u_{jk}$ into a scalar function $f(x)$.
This is a subject of future research.

\section{\label{conc}Conclusion}

The problem of mapping
between Hilbert spaces,
from \emph{IN} of
$\Ket{ \psi}$
to
\emph{OUT} of
$\Ket{\phi}$,
based on a set of wavefunction measurement (within a phase) observations
$\psi_l \to \phi_l$, $l=1\dots M$,
is formulated
as an optimization problem maximizing the total fidelity
$\sum_{l=1}^{M} \omega^{(l)}
\left|\Braket{\phi_l|\mathcal{U}|\psi_l}\right|^2$
subject to probability preservation constraints
on  $\mathcal{U}$.
This optimization problem is reduced to a
novel QCQP problem of maximizing a quadratic form
$\Braket{\mathcal{U}|S|\mathcal{U}}\to\max$
subject to partial unitarity constraints
on $\mathcal{U}$.

The operator $\mathcal{U}$,
represented by a rectangular matrix
$u_{jk}$ of $\dim(OUT) \times \dim(IN)$ dimensions with
$D=\dim(OUT) \le n=\dim(IN)$,
can be viewed as a quantum channel  (\ref{operatorTransform}).
Time evolution represents a special case of this problem.
The optimization problem involving a quadratic function on $\mathcal{U}$
with quadratic constraints is solved using a numerical method outlined
in Section \ref{numericalSolutionSect}.
The method is an iterative approach with a generalized eigenproblem as
its building block.
This differs from commonly used optimization methods
that employ Newtonian iteration or gradient iteration
as their building blocks.
This approach allows us to find the global maximum for almost any input.
In addition, instead of the usual iteration internal state in the form
of a pair (approximation, Lagrange multipliers)
$(u_{jk},\lambda_{ij})$,
the algorithm uses an iteration internal state in the form of a triple:
approximation, Lagrange multipliers, and homogeneous linear constraints
$(u_{jk},\lambda_{ij},C_{d;jk})$.
It is these linear constraints that contribute
to the convergence of the algorithm.

An important feature of the algorithm is that, on each iteration, there is no single solution candidate.
While Newtonian or gradient-based algorithms have a single iterative candidate,
our eigenproblem-based method provides multiple solutions simultaneously (eigenvectors),
the $\lambda_{ij}$ and constraints $C_{d;jk}$ are then calculated from the \hyperref[EVSelectionFromAll]{selected} eigenvector.
Therefore, instead of a single solution, a group of solution candidates (eigenvectors) is obtained at every iteration.
This is similar to genetic programming optimization, where multiple solutions ``flow around'' local maxima and saddle points.
It is the multiple solutions that allow us to identify the global maximum.

This numerical method can be applied to various classical and quantum problems,
such as variational quantum algorithms and quantum mechanical inverse problems,
like recovering a Hamiltonian from a sequence of wavefunction observations.
An operator $\mathcal{U}$ is obtained from these observations,
after which equation (\ref{logUCalc}) can be applied to derive the Hamiltonian.

A demonstration of unitary dynamics $\mathbf{X}^{(l+1)}=\mathcal{U}\mathbf{X}^{(l)}$
for system identification involves determining $\mathcal{U}$
from measured $\widetilde{\mathbf{X}}^{(l)}$,
which are actual $\mathbf{X}^{(l)}$ multiplied by random phase factors.
This was presented for a number of orthogonal operators $\mathcal{U}$
with dimensions $n=D=\{3,5,7,17,40\}$.
An exact recovery of $\mathcal{U}$ was observed.
The technique was also applied to problems involving polynomial bases mapping and scalar function interpolation. An exact solution was obtained for polynomial bases mapping. However, solving the problem of scalar function
interpolation with the developed technique is challenging due to the lack of good equivalence between the problem of mapping between Hilbert spaces
and scalar functions.

The problem of optimal mapping between two Hilbert spaces is reduced to a new algebraic ``eigenoperator'' problem (\ref{eigenvaluesLikeProblem}).
Currently, we only have a numerical solution for it.
An important generalization of Unitary Learning was made in this paper
by considering Hilbert spaces of different dimensions $D\le n$.
A question arises about further generalization.
An important topic for future research could be
\href{https://en.wikipedia.org/wiki/Quantum_operation#Kraus_operators}{Kraus' theorem},
which determines the most general
form of mapping between Hilbert spaces\cite{kraus1983states}:
\begin{align}
  A^{\emph{OUT}}&=\sum\limits_s B_s A^{\emph{IN}} B^{\dagger}_s \label{KrausOperator}
\end{align}
with Kraus operators $B_s$ satisfying
the constraints that the unit $A^{\emph{IN}}$
is converted to the unit $A^{\emph{OUT}}$
\begin{align}
  \sum\limits_s B_sB_s^{\dagger}&=\mathds{1} \label{constraintKrauss}
\end{align}
This is a generalization of the transform (\ref{operatorTransform});
it is applicable
to systems with \href{https://en.wikipedia.org/wiki/Quantum_decoherence}{quantum decoherence}.
In Appendix I of \cite{malyshkin2019radonnikodym}, the corresponding optimization problem is formulated,
but its numerical solution is
the subject of future research\cite{belov2024quantum}.

\begin{acknowledgments}
This research was supported by Autretech Group,
\href{https://xn--80akau1alc.xn--p1ai/}{www.{\fontencoding{T2A}\fontfamily{cmr}\selectfont атретек.рф}},
a resident company of the Skolkovo Technopark.
  We thank our colleagues from the Autretech R\&D department
  who provided insight and expertise that greatly assisted the research.
  Our grateful thanks are also extended
  to Mr. Gennady Belov for his methodological support in doing the data analysis.

V.M. is grateful to Professor Arthur McGurn. Working with him at WMU was an important step in my education, where I learned approaches that combine quantum mechanics and numerical computations.
This paper is dedicated to Professor Arthur McGurn on the occasion of his 75th birthday.

\end{acknowledgments}

\appendix
\section{\label{LambdaSubspace}On Lagrange Multipliers Calculation With Selected States Variation}

In this work, we do not iterate for Lagrange multipliers.
Instead, at each iteration for the current solution approximation $u_{jk}$,
we calculate Lagrange multipliers corresponding to an extremum in this state.
This enhances algorithm stability and allows for directly applying
a solution adjustment procedure (\ref{uAdjEx}) from
the partial constraints (\ref{optimmatrixConstraintScalarAppendix})
to the full set (\ref{optimmatrixConstraintAppendixNUDIAG}).

The variation of the Lagrangian $\mathcal{L}$ (\ref{lagrangetovariateNUDlen})
must be zero for the state $u_{jk}$
\begin{align}
0&=\frac{1}{2}\frac{\delta\mathcal{L}}{\delta u_{iq}}= \sum\limits_{j^{\prime}=0}^{D-1}\sum\limits_{k^{\prime}=0}^{n-1}
S_{iq;j^{\prime}k^{\prime}}u_{j^{\prime}k^{\prime}}
- \sum\limits_{j^{\prime}=0}^{D-1} \lambda_{ij^{\prime}} u_{j^{\prime}q}
\label{variationLagrangiznZero}
\end{align}
These are $Dn$ equations.
The Lagrange multipliers $\lambda_{ij}$
are a Hermitian  $D\times D$ matrix with $D(D+1)/2$
independent elements.
Thus, for a general $u_{jk}$, all $Dn$ equations (\ref{variationLagrangiznZero})
cannot be simultaneously satisfied.
They are satisfied in (\ref{variationUIngivenState})
for the eigenstate $u^{[s]}_{jk}$ of (\ref{EPLeV}),
but after the adjustment (\ref{uAdjEx}), this is no longer the case.
A trivial solution is to take the $L^2$ norm of the variation vector
(\ref{variationLagrangiznZero})
and minimize the sum of squares (\ref{lamMinSQuares})
to obtain (\ref{newLambdaSolPartial}) as the solution to the linear system.
A direct implementation, however, has poor convergence \cite{malyshkin2022machine}.
The idea is to calculate the sum of squares only in specific states.
Consider the problem
\begin{align}
\sum\limits_{s=0}^{N_v-1}
\left|
\sum\limits_{i=0}^{D-1}\sum\limits_{q=0}^{n-1}
v^{[s]}_{iq}\left[
\sum\limits_{j^{\prime}=0}^{D-1}\sum\limits_{k^{\prime}=0}^{n-1}
S_{iq;j^{\prime}k^{\prime}} u_{j^{\prime}k^{\prime}}
\right.\right.
\nonumber \\
\left.\left.
-
\sum\limits_{j=0}^{D-1} \frac{\lambda_{ij}+\lambda_{ji}}{2} u_{jq}
\right]
\right|^2
 \xrightarrow[\lambda_{ij}]{\quad }\min
\label{vSpaceProjL2}
\end{align}
The variation (\ref{variationLagrangiznZero}) is projected onto
$v^{[s]}_{iq}$, $s=0\dots N_v-1$ states,
and the sum of projection squares is taken.
If $v^{[s]}_{iq}$ form a full basis, e.g.,
all the $s=0\dots Dn-1$ eigenvectors $u^{[s]}_{jk}$
of (\ref{EPLeV}), then (\ref{vSpaceProjL2}) is exactly (\ref{lamMinSQuares}).
One may think about (\ref{variationLagrangiznZero}) as the variation of the Lagrangian (\ref{lagrangetovariateNUDlen}), $\delta \mathcal{L}/\delta u_{iq}$.
The sum of squares (\ref{lamMinSQuares})
is $\Braket{\frac{\delta\mathcal{L}}{\delta u}|\frac{\delta\mathcal{L}}{\delta u}}$,
the sum of squares in $v^{[s]}_{iq}$-projected states (\ref{vSpaceProjL2}) is
$\sum_{s=0}^{N_v-1} \Braket{\frac{\delta\mathcal{L}}{\delta u}|v^{[s]}}\Braket{v^{[s]}|\frac{\delta\mathcal{L}}{\delta u}}$, for any full basis $v^{[s]}_{iq}$
they are the same.
In practice, the $v^{[s]}_{iq}$ may not necessarily be full or orthogonal.
If they are a subset of the original eigenstates $u^{[s]}_{jk}$ (\ref{EPLeV})
from the iterative algorithm, they are orthogonal.
If the (\ref{uAdjEx}) adjustment is applied to them, they are not.
One may consider cross states with a $Dn\times Dn$ Gram matrix,
similar to (\ref{psiYlocalized}), as
$\sum_{j,j^{\prime}=0}^{D-1}\sum_{k,k^{\prime}=0}^{n-1}u_{jk} G^{-1}_{jk;j^{\prime}k^{\prime}} v_{j^{\prime}k^{\prime}}$. However, this is usually not necessary since the selection of vectors
$v^{[s]}_{iq}$ is performed solely to improve the algorithm's convergence.

Once the vectors $v^{[s]}_{iq}$  are selected --
in the problem (\ref{vSpaceProjL2}) the $D\times D$ Hermitian matrix $\lambda_{ij}$
should be expressed via  the vector $\lambda_{r}$ of dimension $D(D+1)/2$.
A variation over $\lambda_{r}$ gives a linear system of
dimension $D(D+1)/2$ that can be readily solved.
The number of  $v^{[s]}_{iq}$  vectors should be at least $D(D+1)/2$,
otherwise the linear system will be degenerated.
For matrix to vector conversion and linear system solution
see \texttt{\seqsplit{com/polytechnik/kgo/LagrangeMultipliersPartialSubspace.java:getLambdaForSubspace}},
which implements this functionality
to solve the minimization problem (\ref{vSpaceProjL2}) 
and obtain $\lambda_{ij}$ from a subspace chosen
as the states of high eigenvalues of problem (\ref{EPLeV}).
The success of this approach is moderate: The number of top-$\mu^{[s]}$ states
to select is not precisely clear, the linear system may become degenerate, etc.
One can check these attempts at
\texttt{\seqsplit{com/polytechnik/kgo/LagrangeMultipliersPartialSubspace.java:getLambdaForSubspace}} and their usage in
\texttt{\seqsplit{com/polytechnik/kgo/KGOIterationalLagrangeMultipliersPartialSubspace.java}}. This leads us to conclude
that instead of considering a subspace for constructing  $\lambda_{ij}$
we should consider a subspace for variation of $u_{jk}$.

\subsection{\label{linConst}Linear Constraints On Variation}

Degeneracy of the problem
and quadratic constraints
require
not only a good approximation for Lagrange multipliers,
but also a restricted subspace for variation of $u_{jk}$ in the (\ref{EPL}) problem.
The difficulty arises from partial unitarity
constraints (\ref{optimmatrixConstraintAppendixNUDIAG}).
The optimization (\ref{EPL}) preserves
only the partial constraint (\ref{optimmatrixConstraintScalarAppendix}).
Consider a full orthogonal basis 
$v^{[s]}_{jk}$, where
$\delta_{ss^{\prime}}=\sum\limits_{j=0}^{D-1}\sum\limits_{k=0}^{n-1}v^{[s]}_{jk}v^{[s^{\prime}]}_{jk}$, $s=0\dots Dn-1$,
and a variation vector $\delta u_{jk}$ expanded over this basis
\begin{align}
\delta u_{jk}&=\sum\limits_{s=0}^{Dn-1} a_s v^{[s]}_{jk} \label{uExmansionV}\\
a_s&=\sum\limits_{j=0}^{D-1}\sum\limits_{k=0}^{n-1} v^{[s]}_{jk}\delta u_{jk}
\label{uexpandVCoefs}
\end{align}
If we were working in a regular vector space,
the only available operation would be the scalar product
\begin{align}
\Braket{v|u}&=\sum\limits_{j=0}^{D-1}\sum\limits_{k=0}^{n-1} v_{jk} u_{jk}
\label{scalProductuuD}
\end{align}
Now, when we study partially unitary operators (\ref{operatorTransform}),
we involve a tensor
\begin{align}
G^{v|u}_{ij}&=\sum\limits_{k=0}^{n-1} v_{ik} u_{jk}
\label{tensorVU}
\end{align}
The scalar product corresponds to
$\Braket{v|u}=\mathrm{Tr} G^{v|u}$,
the Gram matrix (\ref{GramUpartial})
is $G^{u}_{ij}=G^{u|u}_{ij}$.
Consider a variation $G^{u+\delta u|u+\delta u}_{ij}$.
To preserve the partial unitarity
constraints (\ref{optimmatrixConstraintAppendixNUDIAG})
on $u_{jk}$ within linear terms (on $\delta u$)
we require all off-diagonal elements of $\mathrm{Herm} G^{u|\delta u}_{ij}$
to be zero $0=G^{u|\delta u}_{ij}+G^{u|\delta u}_{ji}$, $i\ne j$ (homogeneous) and
the diagonal elements to be one (inhomogeneous).
The partial constraint (\ref{optimmatrixConstraintScalarAppendix})
preserves the matrix trace;
thus, it suffices to have the diagonal elements equal,
which forms a homogeneous constraint
$0=G^{u|\delta u}_{ii}-G^{u|\delta u}_{i-1\,i-1}$, $i=1\dots D-1$.
Expanding $\delta u$ in the basis (\ref{uExmansionV}),
we obtain the constraints
\begin{subequations}
\label{subeqConstraint}
\begin{align}
0&=\sum\limits_{s=0}^{Dn-1} a_s \left(G^{u|v^{[s]}}_{ij}+G^{u|v^{[s]}}_{ji}\right)
& \text{$j<i$, $i=0\dots D-1$}
\label{OffDiag0} \\
0&=\sum\limits_{s=0}^{Dn-1} a_s \left(G^{u|v^{[s]}}_{ii}-G^{u|v^{[s]}}_{i-1\,i-1}\right)
& \text{$i=1\dots D-1$}
\label{DiagEq}
\end{align}
\end{subequations}
There are $(D-1)(D+2)/2$ total linear
\href{https://en.wikipedia.org/wiki/System_of_linear_equations#Homogeneous_systems}{homogeneous}
constraints
on the expansion coefficients $a_s$.
Now, not all $Dn$ coefficients $a_s$ are  independent;
there are only $Dn-(D-1)(D+2)/2$ independent ones. 
The constraints (\ref{subeqConstraint}) are homogeneous
\begin{subequations}
\label{constraintsCcoefs}
\begin{align}
 N_d&=(D-1)(D+2)/2 \label{nConstr} \\
 d_{offd}&: 0\dots D(D-1)/2-1 \label{D1C}\\
 d_{diag}&: D(D-1)/2\dots (D-1)(D+2)/2-1 \label{D2C}\\
 s&: 0\dots Dn-1 \\
 C_{d;s}&=
  \begin{cases}
  G^{u|v^{[s]}}_{ij}+G^{u|v^{[s]}}_{ji} & \text{if $d\in d_{offd}$} \\
  G^{u|v^{[s]}}_{ii}-G^{u|v^{[s]}}_{i-1\,i-1} & \text{if $d\in d_{diag}$}
  \end{cases}
\label{constraintsFormula}
\end{align}
\end{subequations}
Here $d\in d_{offd}$ corresponds to $j<i$, $i=0\dots D-1$
and $d\in d_{diag}$ corresponds to $i=1\dots D-1$.
Equations (\ref{constraintsFormula}) can be directly applied to (\ref{linearConstraintsHomog})
after the basis transformation.
\begin{align}
C_{d;jk}&=\sum\limits_{s=0}^{Dn-1} C_{d;s} v^{[s]}_{jk}
\label{CbasisTransform}
\end{align}
There are $D(D-1)/2$ constraints for zero off-diagonal elements (\ref{D1C})
and $D-1$ constraints (one less the dimension)
of  diagonal elements equal to each other (\ref{D2C}).
In total there are  $(D-1)(D+2)/2$ homogeneous constraints.
The constraints $C_{d;jk}$, like the Lagrange multipliers $\lambda_{ij}$,
are calculated solely from the current iteration $u_{jk}$;
see \texttt{\seqsplit{com/polytechnik/kgo/LinearConstraints.java:getOrthogonalOffdiag0DiagEq}} for an implementation.

The result of the consideration above is: if, instead of
a full basis $v^{[s]}_{jk}$ of dimension $Dn$, we take the basis
$V_p$ (\ref{uExprVp}) that has $\mathrm{rank}(C_{d;jk})=(D-1)(D+2)/2$ fewer elements ---
the variation of (\ref{EPLV}) will preserve partial unitarity
of $u_{jk}$
within the first order. This drastically changes the algorithm convergence.
It begins to converge perfectly to a true solution
only if both sets of constraint (\ref{OffDiag0}) and (\ref{DiagEq})
are satisfied;
a single set alone does not ensure convergence.
An iterative algorithm
that finds a partially unitary operator
optimally converting an operator
from the \emph{IN} Hilbert space to the \emph{OUT} Hilbert space (\ref{operatorTransform})
is the main result of this paper.
The result was achieved by considering, on each iteration, not just a pair of approximation and Lagrange multipliers $(u_{jk},\lambda_{ij})$,
but a triple:
approximation, Lagrange multipliers, and homogeneous linear constraints:
$(u_{jk},\lambda_{ij},C_{d;jk})$.
This approach addresses the challenges posed by a quadratically constrained degenerate problem that exhibits local extrema and multiple saddle points.
A similar situation can be observed in dynamic systems with a singular Lagrangian \cite{brown2023singular},
where the solution can be obtained by considering a ``constrained Hamiltonian system''
in which the evolution is constrained to a subspace of the phase space.
In the current work, the constrained subspace (determined by the coefficients $C_{d;jk}$ (\ref{CbasisTransform}))
itself depends on the current iteration $u_{jk}$.

A problem of optimizing the quadratic form
$\sum_{i,j=0}^{D-1} u_i M_{ij} u_j \xrightarrow[u]{\quad }\max$
subject to a single quadratic constraint 
$1=\sum_{i,j=0}^{D-1} u_i Q_{ij} u_j$
where $Q_{ij}$ is a positively definite matrix
can be reduced to an eigenvalue problem.
In Appendix F of \cite{MalMuseScalp}
and later in \cite{boudjemila2021quadratic}
a quadratic form optimization problem with \textsl{two} quadratic constraints
was considered.
An additional constraint was in the form $0=\sum_{i,j=0}^{D-1} u_i C_{ij} u_j$.
This problem is much simpler compared to the one addressed in the current paper.
For this problem, an algorithm using vanilla Lagrange multipliers
iterations converges without requiring additional linear constraints to be added,
see
\texttt{\seqsplit{com/polytechnik/utils/IstatesConditionalV2.java}}
for an implementation.
However, even in this simple case,
it is necessary to try several starting values
for the iterations to find the global maximum. 
Numerical experiments have shown that adding a single linear constraint on
 $u_j$
in each iteration greatly increases the likelihood of finding the global maximum.
This constraint takes the form
$0=\sum_{i,j=0}^{D-1} u_i C_{ij} u^{(cur)}_j$, where $u^{(cur)}_j$
represents the value of $u_j$
to compute the Lagrange multipliers for the current iteration.
A reference implementation
\texttt{\seqsplit{com/polytechnik/utils/IstatesConditionalSubspaceLinearConstraints.java}}
on each iteration solves an eigenproblem of dimension $D-1$
(since a single constraint reduces the dimension by $1$),
always selecting the maximal eigenvalue.
This approach is akin to the heuristic used in
the \hyperref[EVSelectionFromAll]{algorithm described above}.
The result is almost always better than that of
\texttt{\seqsplit{IstatesConditionalV2.java}}
which solves an eigenproblem of dimension $D$
and attempts to select the next iteration from a large number of vectors.
This demonstrates the advantage of considering the iteration state
as a triple (solution, Lagrange multipliers, linear constraints),
even in the simple case of a single additional quadratic constraint.

\section{\label{compComplexityAnalysis}A Preliminary Analysis of Computational Complexity}
Let us estimate the computational complexity of the algorithm. From a sample of $M$ observations (\ref{mlproblemVector}),
the tensor $S_{jk;j^{\prime}k^{\prime}}$ is obtained and used to solve the algebraic problem (\ref{eigenvaluesLikeProblem}).
The calculation of $S_{jk;j^{\prime}k^{\prime}}$, for not very large samples,
contributes little to the overall complexity;
that is, the complexity does not significantly depend on the number of observations in the sample.
The only requirement for the input sample is that it must be information-complete \cite{torlai2023quantum}
in order to recover the $\mathcal{U}$.

The tensor has dimensions $Dn \times Dn$. In each iteration, we solve an eigenvalue problem (\ref{EPLeVV}).
If it were not for the constraints in (\ref{linearConstraintsHomog}),
the dimension of the eigenproblem would be $Dn$. However, the convergence helper constraints from Appendix \ref{linConst}
reduce the dimension to $Dn - (D-1)(D+2)/2$. This is the problem that requires the most computations.
The full list of problems of substantial computational difficulity is as follows:
\begin{itemize}
\item
From $M$ observations (\ref{mlproblemVector}) of vectors with dimensions $n$ and $D$, construct the tensor $S_{jk;j^{\prime}k^{\prime}}$.
The calculations are similar to those used, for example, in covariance matrix calculation.
Each component of the tensor is a sum over all $M$ observations; this task can be trivially parallelized.
\item LU decomposition (\ref{uExprVp}) of the matrix $C_{d;jk}$ (\ref{CbasisTransform}) with dimensions $(D-1)(D+2)/2 \times Dn$.
\item Taking the square root to obtain the matrix $G^{u;-1/2}$ (\ref{mSqrtG})
requires solving an eigenproblem of dimension $D$.
There exist methods to calculate the square root of a positively definite Hermitian matrix
without solving an eigenproblem; see
for example \cite{bjorck1983schur, higham1986newton} and the textbook \cite{bjorck2024numerical}.
Since the dimension of this problem, $D$, is small compared to the dimension
of the eigenproblem (\ref{EPLeVV}) that we consider next,
we conclude that optimizing the square root calculation is not worth the effort.
\item Solving the main eigenproblem (\ref{EPLeVV}), which has a dimension of $Dn - (D-1)(D+2)/2$.
This problem is the most computationally intensive one.
The problem can be parallelized \cite{dongarra1987fully,maschhoff1996p_arpack,katagiri2001efficient},
which can potentially greatly increase the algorithm's performance.
\item Calculating new values of Lagrange multipliers: for the problem addressed in this paper, an analytic solution (\ref{newLambdaSolPartial}) is available. In the general case \cite{belov2024quantum}, a linear system with a dimension equal to the number of independent components in the Lagrange multipliers must be solved. For (\ref{newLambdaSolPartial}), this dimension is $D(D+1)/2$.
\end{itemize}
The problem (\ref{EPLeVV}) is the most computationally difficult. For $D=n$ (unitary learning),
the dimension of this eigenproblem is $N = 1 + n(n-1)/2$. In each iteration,
we need to solve an eigenproblem of dimension $N$.
Finding all eigenvectors has the same complexity as matrix multiplication and is $O(N^3)$ in practice.
However, it can be reduced to $O(N^w)$ for some $2 < w < 3$ \cite{demmel2007fast}.
The state selection \hyperref[EVSelectionFromAll]{step}, however,
typically requires only a single eigenvector corresponding to the maximum eigenvalue.
This problem has lower computational complexity, which can be estimated as $O(N^2)$.
Thus, the algorithm's complexity can be optimistically estimated at $O(n^4)$ in the unitary learning case.

\section{\label{Software}Software description}

\begin{itemize}
\item Install \href{https://www.oracle.com/java/technologies/javase/jdk22-archive-downloads.html}{java 22} or later.
\item Download the latest version of the source code
\href{http://www.ioffe.ru/LNEPS/malyshkin/code_polynomials_quadratures.zip}{\texttt{\seqsplit{code\_polynomials\_quadratures.zip}}}
from \cite{polynomialcode} or from
an \href{https://disk.yandex.ru/d/AtPJ4a8copmZJ?locale=en}{alternative location}.

\item Decompress and recompile the program. Run a simple test
to recover orthogonal matrices of dimensions $3,5,7,17,40$.\\
{\small
\texttt{\seqsplit{unzip\ code\_polynomials\_quadratures.zip}}\\
\texttt{\seqsplit{javac\ -g\ com/polytechnik/*/*java}}\\
\texttt{\seqsplit{java\ com/polytechnik/algorithms/PrintOrthogonalSeq\textbackslash\$TestAuto\ >/tmp/diag\ 2>\&1}}\\
}
The diagnostics is saved to the file \verb+/tmp/diag+
\item Check the maximal absolute difference between the elements
of the original and recovered orthogonal matrices, do \texttt{
  grep DIFF /tmp/diag} \\
\begin{verbatim}
GRAM DIFF for dim=3 is 2.3314683517128287E-15
UNIT DIFF for dim=3 is 4.440892098500626E-16
GRAM DIFF for dim=5 is 6.439293542825908E-15
UNIT DIFF for dim=5 is 7.105427357601002E-15
GRAM DIFF for dim=7 is 5.064698660461886E-14
UNIT DIFF for dim=7 is 1.6431300764452317E-14
GRAM DIFF for dim=17 is 4.6851411639181606E-14
UNIT DIFF for dim=17 is 3.2807090377673376E-14
GRAM DIFF for dim=40 is 4.5630166312093934E-14
UNIT DIFF for dim=40 is 3.907985046680551E-14
\end{verbatim}
Since the unitarity of the test data is exact --- both
quantum channels:
the invariant Gram matrix of Section \ref{UnitaryLearning}
and the invariant unit matrix of Section \ref{traditionalUL}
recover the operator $u_{jk}$ exactly.

\end{itemize}

\bibliography{LD,mla}

\end{document}